\documentclass[10pt,twocolumn,letterpaper]{article}

\usepackage{cvpr}              % To produce the CAMERA-READY version
\usepackage{graphicx}
\usepackage{amsmath}
\usepackage{amssymb}
\usepackage{booktabs}

\usepackage{cite}
\usepackage{indentfirst}
\usepackage{algorithm}
\usepackage{algorithmic}
\usepackage{comment}

\usepackage{xcolor}
\definecolor{cvprblue}{rgb}{0.21,0.49,0.74}
\usepackage[pagebackref,breaklinks,colorlinks,citecolor=cvprblue]{hyperref}
\usepackage[capitalize]{cleveref}
\crefname{section}{Sec.}{Secs.}
\Crefname{section}{Section}{Sections}
\Crefname{table}{Table}{Tables}
\crefname{table}{Tab.}{Tabs.}

\usepackage[labelsep=period]{caption}
\usepackage{enumitem}

\renewcommand{\paragraph}[1]{\vspace{0.2em}\noindent \textbf{#1 \hspace{0.2em}}}

\definecolor{MyDarkRed}{rgb}{0.66, 0.16, 0.16}
\definecolor{MyDarkBlue}{rgb}{0.16, 0.16, 0.66}
\usepackage{lipsum} % Just for generating dummy text, can be removed

\def\cvprPaperID{5881} % *** Enter the CVPR Paper ID here
\def\confName{CVPR}
\def\confYear{2024}

\begin{document}
	
	\title{Robust Overfitting Does Matter: Test-Time Adversarial Purification With FGSM}
	% Test-time Purifying Adversarial Examples with FGSM Robust Overfitting Deep Neural Networks for Adversarial Robust Generalization
	\author{Linyu Tang, Lei Zhang*\\
		School of Microelectronics and Communication Engineering, Chongqing University, China\\
		{\tt\small linyutang@cqu.edu.cn, leizhang@cqu.edu.cn}
	}
	%\and
\maketitle
\label{key}%%%%%%%%% ABSTRACT
\begin{abstract}
	Numerous studies have demonstrated the susceptibility of deep neural networks (DNNs) to subtle adversarial perturbations, prompting the development of many advanced adversarial defense methods aimed at mitigating adversarial attacks. 
	Current defense strategies usually train DNNs for a specific adversarial attack method and can achieve good robustness in defense against this type of adversarial attack.
	Nevertheless, when subjected to evaluations involving unfamiliar attack modalities, empirical evidence reveals a pronounced deterioration in the robustness of DNNs. Meanwhile, there is a trade-off between the classification accuracy of clean examples and adversarial examples. Most defense methods often sacrifice the accuracy of clean examples in order to improve the adversarial robustness of DNNs. To alleviate these problems and enhance the overall robust generalization of DNNs, we propose the \textbf{T}est-Time \textbf{P}ixel-Level \textbf{A}dversarial \textbf{P}urification (TPAP) method. This approach is based on the robust overfitting characteristic of DNNs to the fast gradient sign method (FGSM) on training and test datasets. It utilizes FGSM for adversarial purification, to process images for purifying unknown adversarial perturbations from pixels at testing time in a ``counter changes with changelessness" manner, thereby enhancing the defense capability of DNNs against various unknown adversarial attacks. Extensive experimental results show that our method can effectively improve both overall robust generalization of DNNs, notably over previous methods. Code is available \url{https://github.com/tly18/TPAP}.
\end{abstract}
\section{Introduction}
\label{sec:intro}
Despite the substantial achievements of computer vision tasks such as facial recognition, autonomous driving, and medical image processing, the emergence of adversarial attacks \cite{szegedy2013intriguing} seriously threatens the deployment of computer vision models. Adversarial attacks aim to inject human-imperceptible and malicious noise that are carefully crafted by the adversary into origin clean examples \cite{szegedy2013intriguing}, causing the loss of discriminative capacity in DNNs.\par
\begin{comment}
	\begin{figure}[tb] \centering
		\includegraphics[scale=0.6]{./figure/pur_process.pdf}
		\caption{Our TPAP method in the testing FGSM purification phase process. The black curve indicates the classification boundary, the triangles on the left indicate that the examples belongs to category 1, and the circles on the right indicate that the examples belongs to category 2. The different colors of the graphs indicate that the examples are adversarial examples or purified examples. 
		} \label{fig:figure1}
	\end{figure}
\end{comment}

\begin{figure}[t] \centering\label{fgsm}
	\begin{subfigure}{0.49\linewidth}
		\includegraphics[scale=0.095]{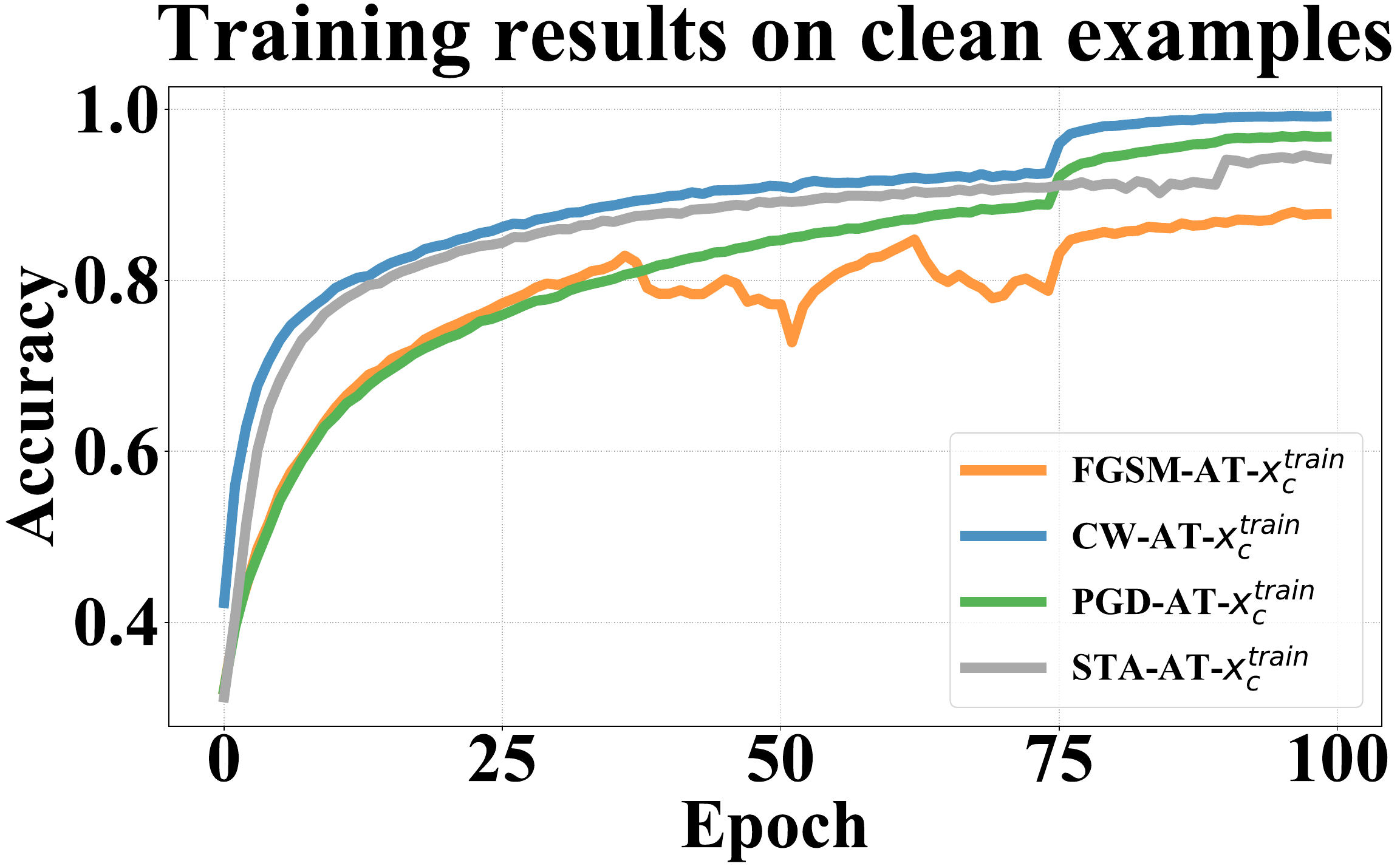}
		\caption{}
		\label{fig:figure1-a}
	\end{subfigure}
	\begin{subfigure}{0.49\linewidth}
		\includegraphics[scale=0.095]{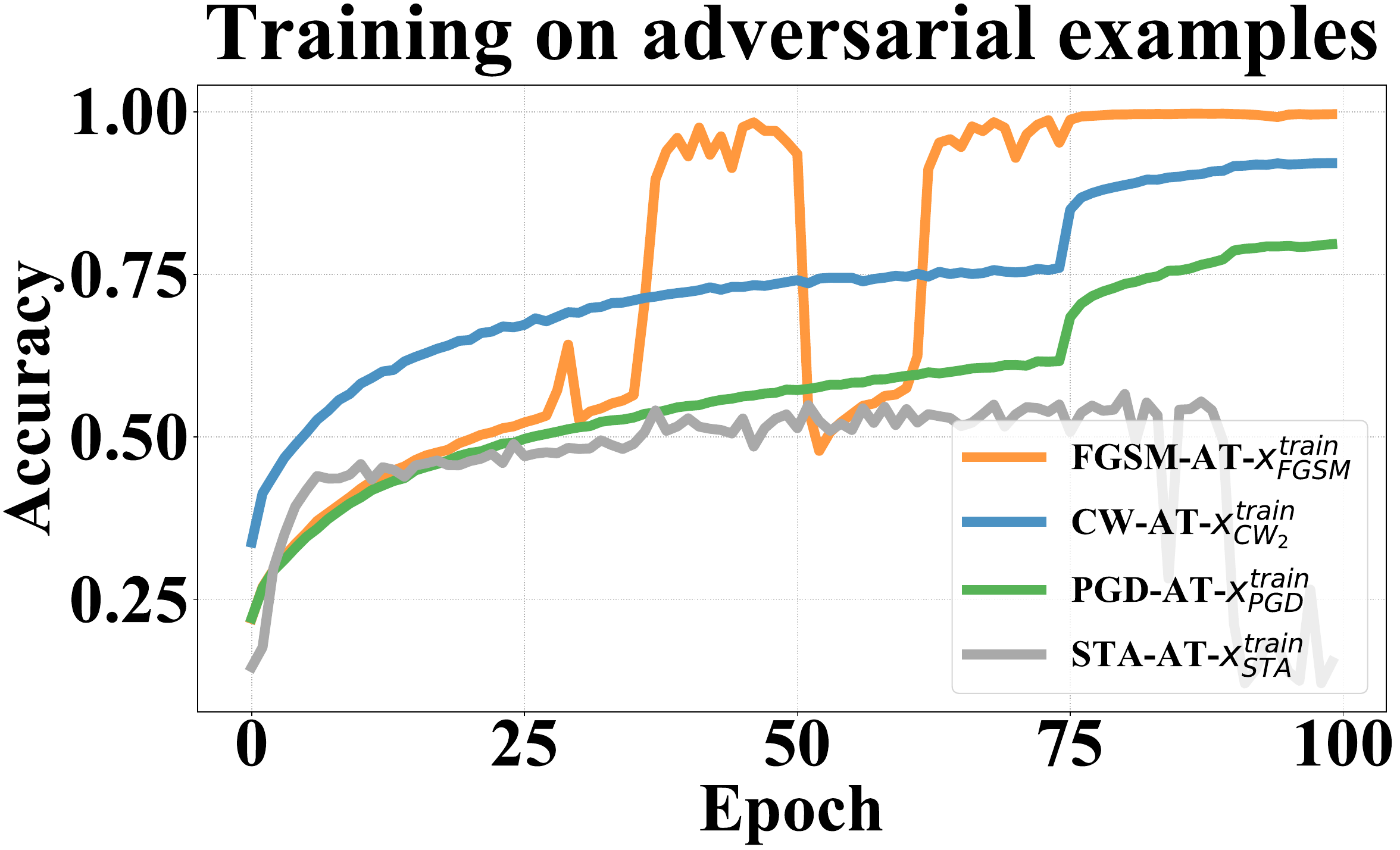}
		\caption{}
		\label{fig:figure1-b}
	\end{subfigure}
	\\
	\begin{subfigure}{0.49\linewidth}
		\includegraphics[scale=0.095]{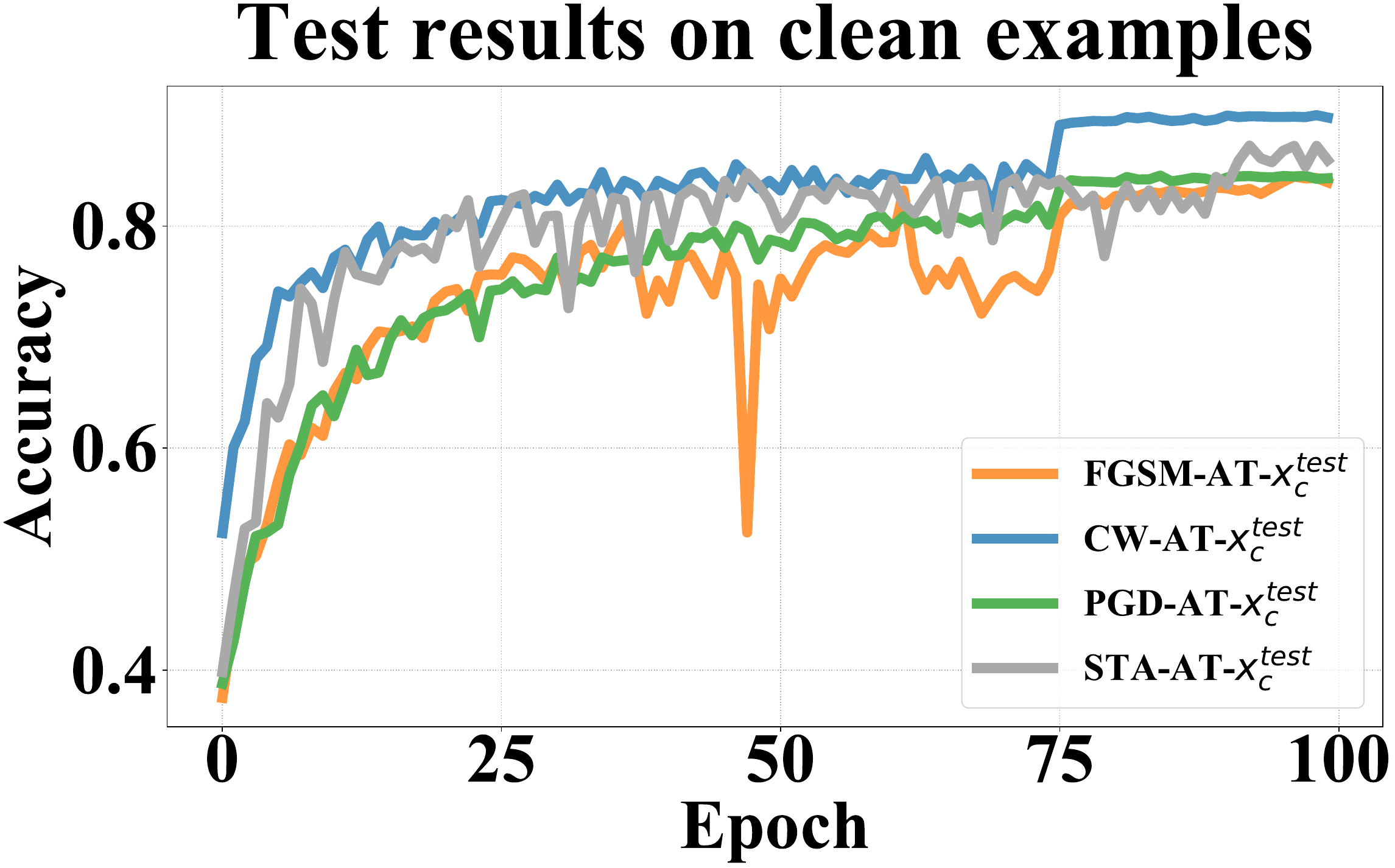}
		\caption{}
		\label{fig:figure1-c}
	\end{subfigure}
	\begin{subfigure}{0.49\linewidth}
		\includegraphics[scale=0.095]{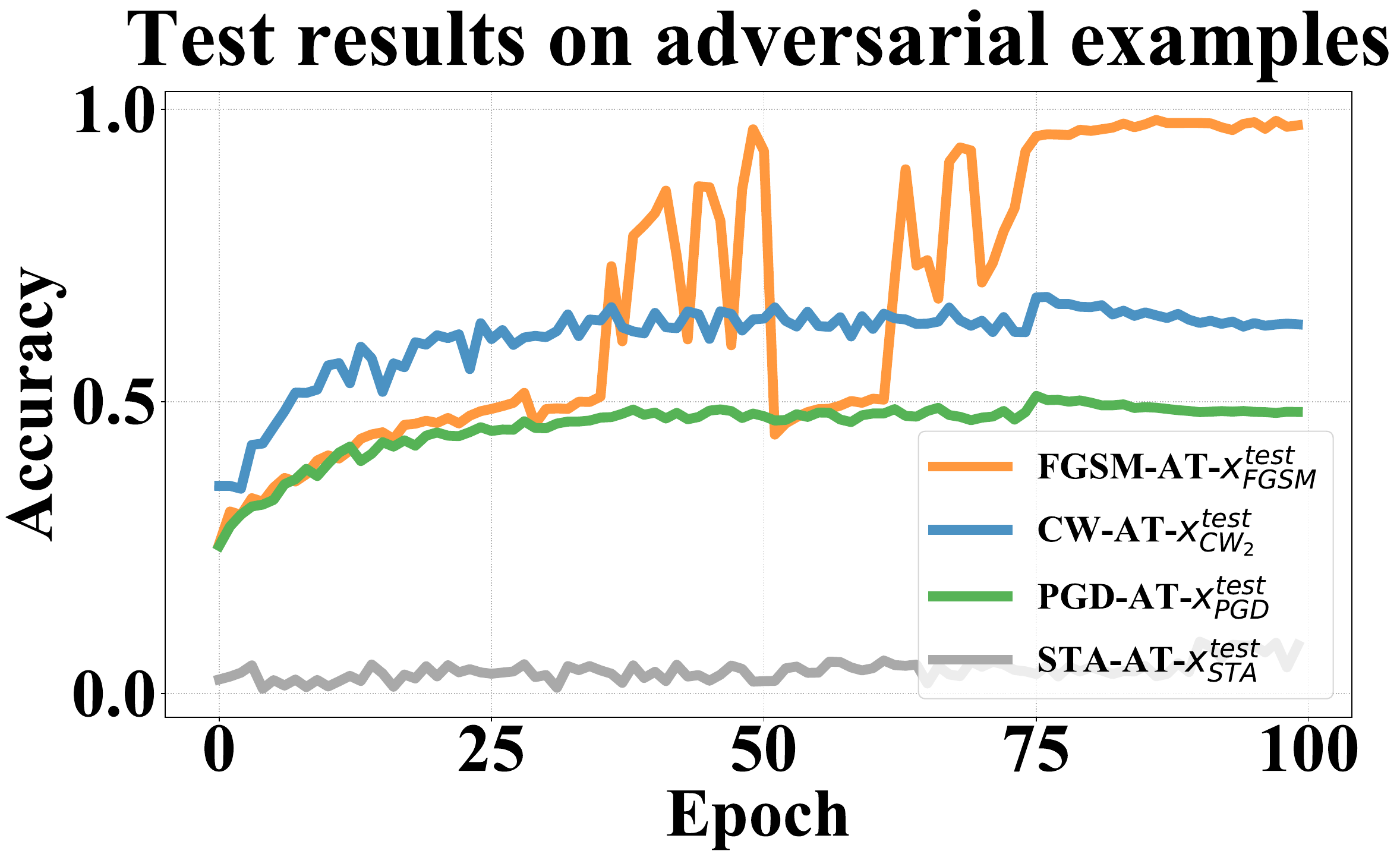}
		\caption{}
		\label{fig:figure1-d}
	\end{subfigure}
	\\
	\caption{
		(a) and (b) represents the training set results of clean and adversarial examples through adversarial training with FGSM, CW$_2$, PGD and STA attacks, respectively, on the ResNet18 using the training set of CIFAR-10, and (c) and (d) show the results of cifar10 test set. The horizontal axis represents epochs, while the vertical axis represents classification accuracy. 
	} 
	\label{fig:figure1}
\end{figure}
In response to adversarial attacks, the concept of adversarial defense was introduced to improve the robustness of DNNs. However, these defense efforts often train DNNs only under the project gradient descent (PGD) \cite{madry2017towards} attack method, without considering whether the networks can defend against other types of attack strategies. 
%In Figure 1, we visualize the differences in the effects of different types of attacks on images. 
As shown in \cref{tab:table1}, our experiment shows a pronounced weakness in defending against attacks such as faster Wasserstein attack (FWA) \cite{wu2020stronger} and spatially transform attack (STA) \cite{xiao2018spatially}. This is easy to understand, because PGD attack is an algorithm reliant on gradient-based technique for crafting adversarial examples, whereas FWA is based on geometrically measured Wasserstein distance in pixel space, and STA introduces method for minimizing spatial deformations through pixel manipulations to generate adversarial examples. Therefore, PGD adversarial training networks cannot learn the defense knowledge of other attack strategies, leading to a serious lack of robust generalization of the networks. In this way, it seems that DNNs are indeed powerless to unknown adversarial attacks?\par
Some intriguing experimental results from prior researches that provide crucial inspiration for the defensive method we proposed in the following sections. Given that PGD adversarial training (PGD-AT) \cite{madry2017towards} is notably time-consuming, Wong \etal~\cite{wong2020fast} introduced a method involving random initialization of fast gradient sign method \cite{goodfellow2014explaining} based adversarial training (FGSM-AT). They observe a phenomenon known as ``catastrophic overfitting" (referred to as FGSM robust overfitting in this paper). Specifically, FGSM-AT networks exhibit classification error rate approaching 100\% when tested on adversarial examples crafted by PGD attack on CIFAR-10 dataset. Andriushchenko \etal~ \cite{andriushchenko2020understanding} conducted the same experiments, further unveiling a noteworthy phenomenon. They observe that FGSM-AT networks achieved an incredibly high rate of correct classification, reaching up to 80\%, when tested on adversarial test examples generated by FGSM attack. 
As shown in \cref{fig:figure1}, we verify the training process of FGSM-AT, CW-AT, PGD-AT, STA-AT that only DNNs trained by FGSM attack reach the robust overfitting state. \cref{fig:figure1-a,fig:figure1-c} show the classification results of the four ATs for clean examples on the training and test sets, respectively, and \cref{fig:figure1-b,fig:figure1-d} show the classification results of the four ATs for adversarial examples generated by their respective adversarial attack methods on the training and test sets, respectively, and only the FGSM-AT has a close to 100\% accuracy on CIFAR-10 for the adversarial examples generated by its own adversarial attack method. 
Since FGSM-AT networks exhibit perfect (overfitting) robustness when dealing with FGSM adversarial examples, this discovery inspires us to explore the possibility that whether the networks can correctly classify unknown adversarial examples after performing FGSM adversarial purification on the testing phase. If successful, this could contribute to enhancing the robust generalization of DNNs. \par
Based on the preceding discussion, we conduct practical experiments and propose the \textbf{T}est-Time \textbf{P}ixel-Level \textbf{A}dversarial \textbf{P}urification (TPAP) method to enhance the robust generalization of DNNs against unknown adversarial attacks.
Specifically, during training phase, we harness the DNN robust overfitting characteristic of FGSM adversarial training to create a network highly adept at classifying clean examples and defending against FGSM attack. 
In the testing phase, images are first fed into the DNN to obtain the pre-predicted labels and their cross-entropy loss which help the input images adapt to the robust overfitting network. These prior knowledge are used to perform FGSM adversarial purification on the image pixels to mitigate their adversarial perturbations, and then classified by the DNN.\par
%For clean examples, if they are correctly classified before purified processing, the classification after processing is equivalent to the classification of adversarial examples with known attack types and adversarial perturbation strength. Our method ensures that the DNN has high classification accuracy on clean data and FGSM adversarial examples trained by FGSM adversarial training, and hence it can correctly classify clean examples with high confidence. 
%If the clean examples are misclassified before processing, it will most likely still be misclassified after processing, but it doesn’t matter.
%For adversarial examples, if they are misclassified before purified processing, the maximum confidence class are the misclassified labels. Using these misclassified labels for FGSM purified processing (untargeted attack) effectively bring the adversarial examples back to the decision boundary, eliminating the adversarial perturbation on the images. \par
%Then, the purified examples are input into the classification network, and the DNN can classify them correctly.\par
%Compared with other methods, our method can further improve the robust generalization of the models on both clean data and adversarial examples. Additionally, our work provides fresh insight into the lack of good generalization of robust models.
Our main contributions are summarized as follows: \par
\begin{itemize}[itemsep=0pt,parsep=0pt,topsep=2bp]
	\item We redefine FGSM robust overfitting deep neural networks (FGSM-RO-DNNs), explore for the first time the effect of hyperparameters on training FGSM-RO-DNNs, and validate the effectiveness of our method on multiple datasets and various DNNs.
	\item Although the adversarial examples are misclassified, they still contain the image semantic information representing their own labels. We propose the TPAP method, which utilizes pre-classification prior knowledge to guide untargeted purification of specific adversarial noise within images, ultimately obtaining correctly classified purified examples.
	\item Our method does not require the extensive use of additional data for training, significantly reducing training time of DNNs and imposing minimal time overhead during testing phase. Empirical experiments show our method presents superior effectiveness against both pixel-constrained and spatially-constrained unseen types of attacks and adaptive attacks, while improving the accuracy of clean examples.
\end{itemize}\par
%The rest of this paper is organized as follows. In Section 2, we briefly review related work on adversarial attacks and adversarial defenses. In Section 3, we describe our defense method and present its implementation. Experimental results against adversarial attacks are provided in Section 4. Finally, we conclude this paper in Section 5.
%-------------------------------------------------------------------------
\section{Related Work}
\label{sec:related_works}
\textbf{Adversarial attack.} 
The adversarial noise generated by the adversarial attack method is limited by a small normball $\begin{Vmatrix}x_a - x_c\end{Vmatrix}_p\leq\epsilon$, which means adversarial examples are similar to their clean examples in perception. Adversarial noise can be crafted by attacking in one or more steps along the direction of the adversarial gradient, such as fast gradient sign method (FGSM) \cite{goodfellow2014explaining}, basic iterative attack (BIA) \cite{kurakin2016adversarial}, momentum iterative attack (MIM) \cite{dong2018boosting}, the strongest first-order information based projected gradient descent (PGD) \cite{madry2017towards}, and the autoattack (AA) \cite{croce2020reliable} method. Furthermore, optimization-based attacks, such as Carlini and Wagner (CW) \cite{carlini2017towards}, decoupling direction and norm (DDN) \cite{rony2019decoupling}, minimize the adversarial noise as part of the optimization objectives. 
The aforementioned attacks directly modify pixel values across the entire image without considering the semantics of the objective, such as shape, outline and posture. These are referred to as pixel-constrained attacks. Furthermore, there are spatial-constrained attacks such as faster Wasserstein attack (FWA) \cite{wu2020stronger}, spatial transform attack (STA) \cite{xiao2018spatially} and robust physical perturbations (RP2) \cite{eykholt2018robust}, which focus on mimicking non-suspicious intentional destructive behavior via geometric structures, spatial transformations or physical modifications.\par
\textbf{Adversarial Defense.} 
Adversarial Training (AT), originally proposed by Ian Goodfellow \etal~\cite{goodfellow2014explaining}, is one of the most classic and effective methods for adversarial defense. Adversarial training refers to the introduction of adversarial examples into training data during network training process to effectively perform data augmentation, so that network learns attack patterns from adversarial examples during training to enhance the robustness.
Research finds that FGSM adversarial training does not always enhance the adversarial robustness of DNNs \cite{moosavi2016deepfool}, as the method of generating adversarial examples through a single linear construction does not evidently produce the optimal adversarial examples.
\cite{li2022subspace,kim2021understanding,wong2020fast,andriushchenko2020understanding, huang2023fast, jia2023improving, li2020towards,vivek2019regularizer} focused on improving FGSM adversarial training to prevent catastrophic overfitting.
Madry \etal~\cite{madry2017towards} proposed using a stronger PGD attack for adversarial training. They formulated adversarial training as a min-max optimization problem and demonstrated PGD adversarial training can obtain a robust network.
\cite{tsipras2018robustness, madry2017towards,zhang2019theoretically} showed that there is a negative correlation between the clean accuracy and adversarial robustness of DNNs. 
Therefore, Zhang \etal~ proposed TRADES \cite{zhang2019theoretically} to decompose the prediction error for adversarial examples (robust error) as the sum of the natural (classification) error and boundary error. They design a new training objective function to balance adversarial robustness and natural accuracy.
Wang \etal~\cite{wang2019improving} found that misclassified clean examples have a significant impact on the final robustness of DNNs. They proposed the MART algorithm, which explicitly distinguishes between misclassified and correctly classified examples during training, thus constraining misclassified clean examples using a weighting coefficient on loss function to significantly improve the adversarial robustness of the network.
Wei \etal~\cite{wei2023cfa} studied the preferences of different categories for adversarial perturbations and introduced a category-calibrated fair adversarial training framework that automatically tailors specific training configurations for each category.
To alleviate these conflicted dynamics of the decision boundary, Xu \etal~\cite{xu2023exploring} proposed Dynamics-Aware Robust Training (DyART), which encourages the decision boundary to engage in movement that prioritizes increasing smaller margins.
Wang \etal~\cite{wang2023better} proposed to exploit better diffusion networks to generate much extra high-quality data for adversarial training, which can improve the robustness accuracy of DNNs.
\cite{wu2020adversarial,sriramanan2020guided,sriramanan2021towards,zhang2020attacks} were dedicated to improving training methods and training loss functions to enhance the robustness of DNNs.\par
\textbf{Adversarial Purification.} 
The goals of both adversarial purification and adversarial training are to enhance the resilience of DNNs against adversarial attacks. Adversarial training primarily focuses on improving robustness through network training, while adversarial purification places emphasis on purifying input data before feeding it into the classification network during testing to mitigate the impact of adversarial perturbation.
\cite{akhtar2018defense,zhou2021towards,zhou2021removing,yuan2020ensemble,yang2021class,lin2020dual} added additional network for image purification, such as VAE \cite{kingma2013auto,hwang2019puvae}, GAN \cite{goodfellow2014generative,samangouei2018defense}, DUNET \cite{liao2018defense}, and then jointly trained with the classification network to make the classification results and image pixel values of the adversarial examples consistent with the clean examples. However, some of these methods would reduce the accuracy of clean examples and some are computationally intensive. Testing phase adversarial purification is the process of converting adversarial examples encountered by the network in actual inference into clean data representations. %after the network has been trained. 
Shi \etal~ \cite{shi2021online} introduced SOAP, which employs double consistency losses during the training phase and self-supervised loss during testing phase to purify adversarial examples. SOAP provides flexibility against adversarial attacks compared to networks that use a fixed architecture during testing. \cite{wang2022guided,nie2022diffusion} used diffusion network for denoising adversarial examples, where Gaussian noises are gradually added to submerge the adversarial perturbations during the forward diffusion process and both of these noises can be simultaneously removed following a reverse reconstruction process.
%Combined with massive data enhancement, the robustness of DNNs is greatly enhanced. 
However, these methods need long training time and large memory.
%-------------------------------------------------------------------------
\begin{comment}
\end{comment}
\begin{figure}[tb] \centering\label{method}
	\includegraphics[scale=0.4]{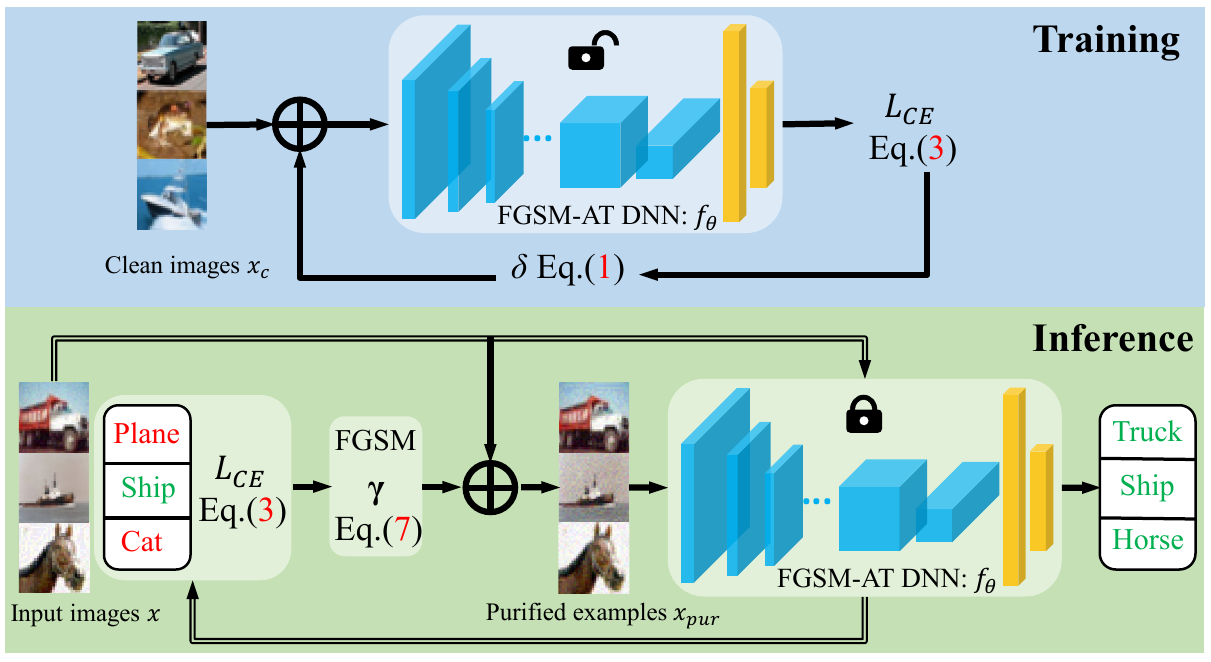}
	\caption{Overview of the training phase and testing phase inference phase. Arrows indicate data flow, and double straight arrows indicate testing phase pre-classification.
		%In the training phase, the network is trained using FGSM adversarial examples to achieve FGSM robust overfitting. In the testing phase, input examples, including clean examples and adversarial examples, are processed using FGSM adversarial purification to obtain purified examples, which are then fed into the DNN classifier.
	} \label{fig:figure2}
\end{figure}
\section{Method}%
\label{sec:method}
This paper investigates how to achieve robust DNNs by utilizing the under-studied FGSM robust overfitting prior. We present the \textbf{T}est-Time \textbf{P}ixel-Level \textbf{A}dversarial \textbf{P}urification (TPAP) method, a novel defense strategy that uses a FGSM robust overfitting network and adversarial purification processing at testing phase for robust defense against unknown adversarial attacks, as illustrated in \cref{fig:figure2}.
\subsection{Preliminary}%
\label{sub:Preliminary}
This paper primarily focuses on the task of image classification under adversarial attacks. We use $x_c$ to represent clean examples, $x_a$ for adversarial examples, $x$ for input images including clean examples $x_c$ and adversarial examples $x_a$, $x_{pur}$ for purified examples, $y$ for the true labels corresponding to the images, and $y_{pred}(\cdot)$ for the prediction labels. $\delta$ represents the adversarial perturbation added to the image pixels when generating adversarial examples and $\gamma$ represents the adversarial purification applied to the image pixels during the test-time purification phase. $\epsilon$ and $\xi$ respectively denote the maximum magnitude of pixel value changes in the generated adversarial examples and the purified images at testing phase. $\alpha$ and $\beta$ denote the step size. $C$ represents the number of categories in the dataset. we use $\theta$ to denote the weight parameters of a DNN $f$.
\subsection{Robust Overfitting Prior of FGSM-AT}%
In this paper, we redefine the network robust overfitting as follows: on the training set, the classification accuracy of adversarial examples generated by a specific and trained attack method is higher than 90\%, and the classification accuracy of other kinds of adversarial examples is less than 10\%. Most importantly, on the test set, the classification accuracy of clean examples and adversarial examples crafted through known attack on DNNs is high. This FGSM robust overfitting is shown after 80 epochs in \cref{fig:figure3-a,fig:figure3-b}.\par 
During the training phase of DNN, the network undergoes adversarial training using FGSM adversarial examples until DNN reaches the state of FGSM robust overfitting. Formally, FGSM adversarial examples with $\ell_\infty$-norm for $\alpha=\epsilon$ are computed by,
\begin{equation}\label{adver}
	\delta=\alpha*sign({\frac{\partial{L_{CE}(f_\theta(x_c),y)}}{\partial{x_c}}})
	%\rm s.t. \quad ||\delta||_p\leq\epsilon
\end{equation}
\begin{equation}\label{limit_c}
	x_a=x_c+\delta%max(min(\delta,\epsilon),-\epsilon)
\end{equation}
where $L_{CE}$ represents cross-entropy loss defined as, 
\begin{equation}
	\begin{split}
		L_{CE}(f_\theta(x),y)&=-{\sum_{s=0}^{C-1} P(x,s)* log{\frac{e^{f_\theta(x,s)}}{\sum_{k=0}^{C-1} e^{f_\theta(x,k)}}}}	\\
		&=-log{\frac{e^{f_\theta(x,y)}}{\sum_{k=0}^{C-1} e^{f_\theta(x,k)}}}
	\end{split}
\end{equation}
where $P(x,s)\in \mathbb{R}^1$ denotes the probability categorized as $s$ in the ground-truth label $P(x)\in \mathbb{R}^C$.
$f_\theta(x)\in \mathbb{R}^C$ is the output of DNN, $f_\theta(x,k) \in \mathbb{R}^1$ represents the value of the output neuron $k$.
The partial derivatives of the cross-entropy loss function with respect to the network weights are calculated and updated as,%, updating the network representation as,
%Calculate the partial derivatives of the cross-entropy loss function with respect to the network weights, updating the network representation as,
\begin{equation}\label{update}
	\theta=\theta-{\frac{\partial{L_{CE}(f_\theta(x_a),y)}}{\partial{\theta}}}
\end{equation}
\subsection{Test-time Pixel-Level Purification}%
During the test phase, the parameters of the FGSM robust overfitting network (FGSM-RO-DNN) obtained during the training phase are frozen.
We expect to start from the pixels of images, purifying adversarial examples to ensure accurate classification without affecting the correct classification results of the purified clean examples. Our idea is expressed by the following equation, 
\begin{equation}
	%\gamma = \arg\min_{||\gamma||_p \leq \xi} L_{\text{CE}}(f_{\theta}(x + \gamma), \text{Label}(f_{\theta}(x)))	\\
	\gamma = \operatorname*{argmin}_{||\gamma||_p \leq \xi} L_{\text{CE}}(f_{\theta}(x + \gamma), \text{Label}(f_{\theta}(x)))
\end{equation}
\indent In the specific implementation, the images containing clean and adversarial examples are directly fed into the FGSM-RO-DNN to obtain pre-predicted labels and the partial derivatives of their cross-entropy loss function for each pixel on the input images.
We choose the FGSM robust overfitting network based on 2 considerations: 1) it is highly robust to pairs of clean examples and FGSM adversarial examples and 2) is not resistant to other unknown types of attacks. In other words, FGSM-RO-DNN is more suitable for the purification process in the testing phase than other networks because even though images are easily attacked by other types of attack methods, they can be corrected in FGSM adversarial purification.
%images are easily corrected in FGSM adversarial purification although they are very easy to attack down by other types of attack methods. %The robust overfitting network is very vulnerable to other types of attacks but very robust to FGSM attack, so it is suitable for the purification process in the testing phase than other networks. \par
\begin{figure}[t] \centering
	\begin{subfigure}{0.49\linewidth}
		\includegraphics[scale=0.330]{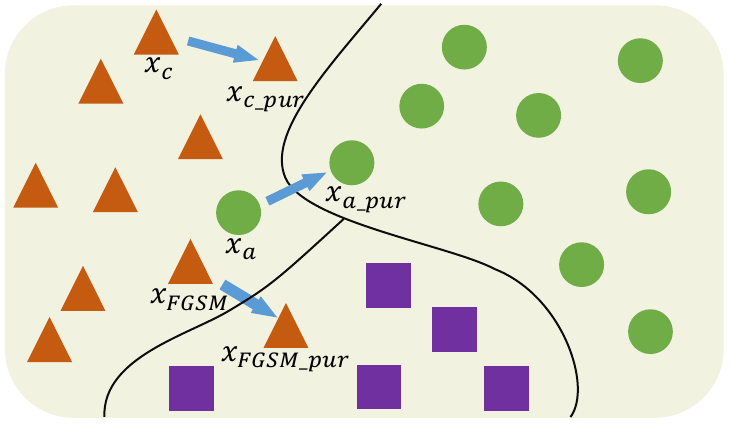}
		\caption{}
		\label{fig:figure7-a}
	\end{subfigure}
	\begin{subfigure}{0.49\linewidth}
		\includegraphics[scale=0.330]{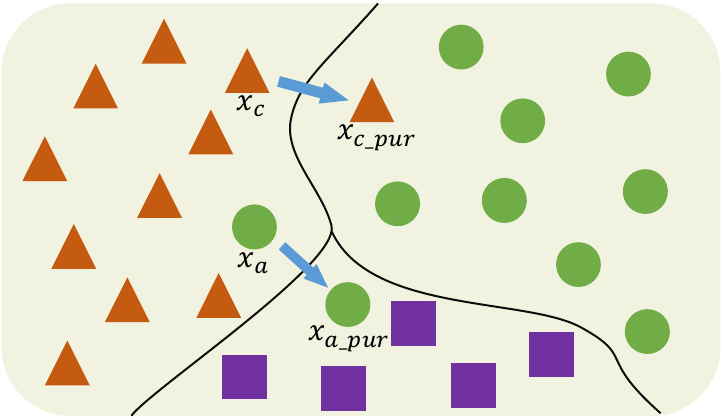}
		\caption{}
		\label{fig:figure7-b}
	\end{subfigure}
	\\
	\caption{
		(a) and (b) represent the processing of FGSM robust overfitting and other adversarial training methods for DNNs in the test purification phase, respectively. The black curves indicate the categorization boundaries, and the triangles, circles, and squares indicate the 3 different categories, respectively.
	} 
	\label{fig:figure7}
\end{figure}
As shown in \cref{fig:figure7-a}, after purification, it is able to classify the $x_c$ and $x_a$ correctly, however \cref{fig:figure7-b} is not. Our experiments in \cref{tab:table7} validate the experimental results using other attack methods instead of FGSM.
As shown in \cref{fig:figure7-a}, for clean examples, if they are correctly classified before purification processing, the classification after processing is amount to the classification of adversarial examples with a known attack type. %and adversarial perturbation strength
Our method ensures that FGSM-RO-DNN has high classification accuracy on clean data and FGSM adversarial examples, and hence it can correctly classify clean examples with high confidence. 
If clean examples are misclassified before purification, the likelihood of  misclassification after purification is high, but and negligible due to its low possibility.
For adversarial examples, if they are misclassified before purification processing, maximum confidence classification results are misclassified labels. Using these misclassified labels for FGSM untargeted adversarial purification effectively bring adversarial examples back to the decision boundary, eliminating adversarial perturbation. Then, purified examples are fed into the classification network, and FGSM-RO-DNN can correctly classify them due to its robustness to FGSM attcak.

In adversarial purification process, it is necessary to satisfy pixel change constraints and ensure that the image semantic information is not disrupted.
Formally, the output of the DNN w.r.t. input $x$ is represented as $f_\theta(x)=a_0, a_2, …, a_{c-1}$. The predicted label is the position of the largest neuron denoted as,
\begin{equation}
	\begin{split}
		\label{pre-predict}
		y_{pred}(x)&=\text{Position}(\text{max}(a_0, a_2, …, a_{c-1}))	
		\\
		&=\text{Label}(f_\theta(x))
	\end{split}
\end{equation}
\indent The FGSM purification implementation with $\ell_\infty$-norm for $\beta=\xi$ is computed by,
\begin{equation}\label{purify}
	\gamma=\beta*sign({\frac{\partial{L_{CE}(f_\theta(x),y_{pred}(x))}}{\partial{x}}} )
\end{equation}
\begin{equation}\label{limit_a}
	x_{pur}=x+\gamma%max(min(\gamma,\xi),-\xi)
	%			s.t.||\gamma||_p\leq\xi
\end{equation}
\indent Finally, the purified example is fed into the network again to obtain the final classification prediction results.
\begin{equation}\label{result}
	y_{pred}(x_{pur})=\text{Label}(f_\theta(x_{pur}))
\end{equation}
\indent The algorithm of TPAP is summarized in Algorithm \ref{alg:algorithm-label}.
\begin{algorithm}[h]
	\small
	\caption{: The Algorithm of TPAP.}
	\label{alg:algorithm-label}
	\vspace{5pt}
	\textbf{Training phase: FGSM-RO-DNN training}
	\vspace{1pt}
	\hrule % 横线
	\vspace{5pt}	
	\begin{algorithmic}[1]
		\STATE \textbf{Input:}  Training set $x_c$, network $f_\theta(\cdot)$ parameterized by $\theta$, batch size $B$, perturbation radius $\epsilon$, total number of iterations $epochs$;
		\STATE \textbf{Output:} Robust overfitting network trained with TPAP.
		% 调整\abovedisplayskip和\belowdisplayskip的值以减小间距
		\setlength{\abovedisplayskip}{1pt}
		\setlength{\belowdisplayskip}{1pt}
		
		\FOR{$i = 1$ to $epochs$}
		\FOR{$j = 1$ to $B$ (in parallel)}
		\STATE
		Obtain FGSM adversarial perturbations using \cref{adver}. 
		\STATE Obtain FGSM adversarial examples $x_{a}$ using \cref{limit_c}.
		\STATE Update network weights with the optimizer in \cref{update}.
		\ENDFOR
		\ENDFOR
		\RETURN the network weight parameters $\theta$.
	\end{algorithmic}
	\vspace{1pt}
	\hrule % 横线
	\vspace{8pt}
	\textbf{Testing phase: Test-time purification processing}
	\vspace{1pt}
	\hrule % 横线
	\vspace{5pt} 	
	\begin{algorithmic}[1]
		\STATE\textbf{Input:} Test set $x$ including clean examples $x_c$ and adversarial examples $x_a$, pre-trained robust overfitting network $f_\theta(\cdot)$, purification radius $\xi$;
		\STATE\textbf{Output:} Prediction labels $y_{pred}$.
		% 调整\abovedisplayskip和\belowdisplayskip的值以减小间距
		\setlength{\abovedisplayskip}{0pt}
		\setlength{\belowdisplayskip}{0pt}
		\FOR{$i = 1$ to $B$ (in parallel)}
		\STATE Compute the pre-prediction of input images using \cref{pre-predict}.
		\STATE Perform adversarial purification using \cref{purify}.
		\STATE Obtain purified examples $x_{pur}$ using \cref{limit_a}.
		\STATE Compute the output of $x_{pur}$ and obtain the maximum classification results as their final prediction labels in \cref{result}.
		\ENDFOR
		\RETURN $y_{pred}(x_{pur})$
	\end{algorithmic}
\end{algorithm}
\begin{table*}[t] \centering
	\caption{Classification accuracy rates (percentage) against white attacks on ResNet-18, VGG-16, WideResNet-34 for CIFAR-10, CIFAR-100, SVHN and Tiny-ImageNet datasets. The best results are highlighted in \textbf{bold} and the second best in \underline{underline}.}
	\label{tab:table1}
	\newcommand{\Natural}{Clean\xspace}
	\newcommand{\FGSM}{FGSM\xspace}
	\newcommand{\PGDT}{PGD-20\xspace}
	\newcommand{\PGDO}{PGD-100\xspace}
	\newcommand{\CW}{CW$_2$\xspace}
	\newcommand{\DDN}{DDN$_2$\xspace}
	\newcommand{\AAA}{AA\xspace}
	\newcommand{\STA}{STA\xspace}
	\newcommand{\FWA}{FWA\xspace}
	\newcommand{\DIM}{TI-DIM\xspace}
	
	\makebox[\textwidth]{\large} 
	\resizebox{\textwidth}{!}{
		\large
		\begin{tabular}{l|*{10}{c}|*{10}{c}}
			\toprule
			\fontsize{15}{20}\selectfont ResNet-18 & \multicolumn{10}{c|}{\fontsize{15}{20}\selectfont CIFAR-10} & \multicolumn{10}{c}{\fontsize{15}{20}\selectfont CIFAR-100}\\ 
			\midrule
			Method & \Natural & \FGSM & \PGDT & \PGDO  & \CW & \DDN & \AAA & \STA & \FWA & \DIM & \Natural & \FGSM & \PGDT & \PGDO  & \CW & \DDN & \AAA & \STA & \FWA & \DIM \\
			\midrule
			PGD-AT\cite{madry2017towards}	 	 & \underline{84.54}& 55.11 	& 48.91 		& 47.7 			& 59.15 		& 19.15 		& 43.3 			& 0.35 	      & 3.19 		& 49.23 				& \underline{57.77}  & 28.68 		& 25.52 		& 24.99 		& 30.42 		& 11.03 		& 21.21 		& 0.03 		    & 2.55 			 & 25.79\\
			TRADES\cite{zhang2019theoretically}   	 & 83.22		 & 58.51 		& 54.97 		& 54.07 		& 71.62 		& 24.24 		& 48.97 		& 1.09 		  & 5.38 		& 55.14 				& 53.93 		& 30.22 		& 28.06 		& 27.77 		& 35.35 		& 14.16 		& 23.01 		& 0 		    & 5.25 	 		 & \underline{28.25}\\
			MART\cite{wang2019improving} 		 & 82.14 	& \underline{59.57} & 55.39 		& 54.8 			& 74.3 			& 25.83 		& 47.84 		& 2.06 		  & 6.43 		& \underline{56.24} 			& 55.52 		& 30.83 		& 28.38 		& 28.16 		& 35.57 		& 13.78 		& 23.01 		& 0.02 		    & 3.44 		 	 & \textbf{28.51}\\
			SOAP\cite{shi2021online}         & 84.07         & 51.02		& 51.42			& -				& 73.95			&	-			&-				&-			  &	-			&-						&52.91 & 22.93 &27.55 & -  & 50.26 &-  &-&-&-&-\\
			\midrule
			TPAP(Ours) 	 &\textbf{86.25}	&\textbf{61.41}	&\textbf{79.06}	&\textbf{80.5}	&61.37	&64.5	&\textbf{76.34}	&31.4	&\textbf{52.83}	&\textbf{75.21}
			& 57.43 		& \textbf{35.64}& \underline{44.69}  &\textbf{42.23} & 48.7 			& 50.84  		&\textbf{47.48} & 43.8 		    &\textbf{32.23}  & 27.84\\
			TPAP+TRADES  & 84.07 		 & 44.16 		& 73.02 		& 66.12 & \underline{90.87}  & \underline{87.29}& \underline{74.94}  & \underline{80.38} & \textbf{51.34} & 31.52 			& 57.67 		& 27.71 		& 37.82 		& 32.93 		& \textbf{70.49}&\textbf{65.62} & 35.23 		&\textbf{66.92} & 27.06 	     & 15.41 \\
			TPAP+MART    & 84.06		 & 43.6 & \underline{73.69}  & \underline{69.78}& \textbf{92.38}& \textbf{90.05}& 72.11 		& \textbf{85.7} & 46.69 		& 23.25 			& \textbf{61.03}& \underline{32.6} 	&\textbf{44.9} 	& \underline{39.49} 	& \underline{68.61}  & \underline{61.38} 	& \underline{46.19} 	& \underline{66.39}  & \underline{28.45} 	 & 15.66\\		
			\bottomrule
		\end{tabular}
	}
	
	\vspace{3pt}
	\makebox[\textwidth]{\large} 
	\resizebox{\textwidth}{!}{
		\large
		\begin{tabular}{l|*{10}{c}|*{10}{c}}
			\toprule
			\fontsize{15}{20}\selectfont ResNet-18 & \multicolumn{10}{c|}{\fontsize{15}{20}\selectfont SVHN} & \multicolumn{10}{c}{\fontsize{15}{20}\selectfont Tiny-ImageNet}\\ 
			\midrule
			Method & \Natural & \FGSM & \PGDT & \PGDO  & \CW & \DDN & \AAA & \STA & \FWA & \DIM & \Natural & \FGSM & \PGDT & \PGDO  & \CW & \DDN & \AAA & \STA & \FWA & \DIM \\
			\midrule
			PGD-AT\cite{madry2017towards} 		& 91.66 		&  \textbf{87.93}& 63.86 		& 44.57 		& 72.65 			& 8.84 			& 31.35 			& 6.51    		& 10.1 			& \textbf{68.19} 		& \textbf{49.06} 	& 24.26  			& 22.09 		& 21.44				& 28 		 		& 30.41 		& 16.82 			& 0.21    			& 0.99 			& 22.06\\
			TRADES\cite{zhang2019theoretically} 		& 91.32 		& 73.38    		& 59.01 		& 56.31 		& 72.96 			& 5.19 			& 47.03 			& 1.57 			& 0.31 			& 59.18 				& 46.59 			& 22.9    			& 21.46 		& 21.03 			& 28.87 			& 28.8 			& 15.99 			& 0 				& 1.8 			& 21.54\\
			MART\cite{wang2019improving}	&\underline{91.81}&\underline{75.31}& 56.55 		& 51.28 		& 71.45 			& 6.96 			& 42.08 			& 0.9 			& 0.86 			& 56.68 				& 46.21  			& \underline{25.73} & 24.16 		& 23.59 			& 30.58 			& 30.24 		& 17.85 			& 0.34 				& 1.75 			& 24.23\\
			\midrule
			TPAP(Ours) 	& 89.62 		& 67.56 		& \textbf{83.62}& \textbf{85.25}& 51.39 			& 62.07 		& \textbf{88.76}	& 55.12 		& \textbf{60.56}& 40.99 				& 48.72 			& \textbf{46.6} 	& \underline{37.88}& \textbf{36.87} 	& {31.48} & \underline{45.28}& \textbf{39.8}   	& {12.43} & \textbf{38.31}  & \textbf{32.93}\\
			TPAP+TRADES & 91.36 		& 41.22 		& 80.31 	  &\underline{72.77}& \underline{92.14} & \textbf{88.96}& \underline{66.57} & \underline{88.31}& 28.67 		& \underline{59.64} 	& 46.22 		&8.96
			&17.93  	&14.07	&\textbf{48.54}	 &\textbf{47.95}	&20.5	&\textbf{42.48}	&5.4	&5.5	\\  	
			TPAP+MART 	& \textbf{93.74}& 26.92 	& \underline{81.62} & 66.31			& \textbf{93.28}  &\underline{88.36}& 63.26 		& \textbf{90.06}& \underline{28.72} & 26.65 				& \underline{48.88} & 18.46 			&\textbf{38.79}  & \underline{36.53} & \underline{41.23} & {43.42}& \underline{31.72} & \underline{26.81}  	& \underline{36.49}  & \underline{26.93}\\
			\bottomrule
		\end{tabular}
	}
	
	\vspace{3pt}
	\makebox[\textwidth]{\large} 
	\resizebox{\textwidth}{!}{
		\large
		\begin{tabular}{l|*{10}{c}|*{10}{c}}
			\toprule
			\fontsize{15}{20}\selectfont VGG-16 & \multicolumn{10}{c|}{\fontsize{15}{20}\selectfont CIFAR-10} & \multicolumn{10}{c}{\fontsize{15}{20}\selectfont CIFAR-100}\\ 
			\midrule
			Method & \Natural & \FGSM & \PGDT & \PGDO  & \CW & \DDN & \AAA & \STA & \FWA & \DIM & \Natural & \FGSM & \PGDT & \PGDO  & \CW & \DDN & \AAA & \STA & \FWA & \DIM \\
			\midrule
			PGD-AT\cite{madry2017towards} &81.11 &51.91 &45.18 &43.78 &56.12 &24.75 &39.26 &0.26 &4.55 &45.51 &50.68 &24.4 &20.87 &20.31 &25.27 &11.9 &17.84 &0 &3.29 &21.1	\\
			TRADES\cite{zhang2019theoretically} &78.75 &52.84 &49.23 &48.21 &66.68 &29.03 &43.01 &1.02 &6.79 &49.41 &48.41 &\textbf{26.77} &24.52 &24.19 &31.25 &14.51 &20.23 &0 &6.52 &24.72\\
			MART\cite{wang2019improving}	 &77.79 &\textbf{54.03} &50.42 &49.57 &67.47 &29.48 &43.4 &1.75 &6.04 &50.72 &49.05 &\underline{25.46} &23.02 &22.53 &28.4 &12.53 &19.3 &0.07 &4.78 &23.12\\
			\midrule
			TPAP(Ours) &77.02 &\underline{53.4} &63.99 &57.68 &40.57 &31.2 &38.71 &\textbf{89.5} &29.29 &\underline{54.65} &48.11 &5.93 &\underline{55.25} &\underline{54.9} &26.8 &16.9 &29.65 &5.73 &7.16 &\textbf{44.76}	\\
			TPAP+TRADES &\textbf{89.13} &41.55 &\underline{78.13} &\underline{74.01} &\textbf{88.98} &\textbf{87.38} &\underline{74.64} &\underline{61.17} &\textbf{31.48} &48.09	&\underline{59.85} &25.36 &49.04 &42.99 &\textbf{61.15} &\textbf{55.66} &\underline{47.71} &\textbf{53.18} &\underline{26.54} &{22.9}\\  	
			TPAP+MART &\underline{88.1} &23.44 &\textbf{89.98} &\textbf{88.42} &\underline{83.51} &\underline{78.37} &\textbf{85.08} &51.52 &\underline{31.13} &\textbf{56.77}	&\textbf{62.04} &19.41 &\textbf{61.79} &\textbf{58.28} &\underline{59} &\textbf{51.45} &\textbf{50.23} &\underline{35.41} &\textbf{29.67} &\underline{31.82}\\
			\bottomrule
		\end{tabular}
	}
	\vspace{3pt}
	\makebox[\textwidth]{\large} 
	\resizebox{\textwidth}{!}{
		\large
		\begin{tabular}{l|*{10}{c}|*{10}{c}}
			\toprule
			\fontsize{15}{20}\selectfont VGG-16 & \multicolumn{10}{c|}{\fontsize{15}{20}\selectfont SVHN} & \multicolumn{10}{c}{\fontsize{15}{20}\selectfont Tiny-ImageNet}\\ 
			\midrule
			Method & \Natural & \FGSM & \PGDT & \PGDO  & \CW & \DDN & \AAA & \STA & \FWA & \DIM & \Natural & \FGSM & \PGDT & \PGDO  & \CW & \DDN & \AAA & \STA & \FWA & \DIM \\
			\midrule
			PGD-AT\cite{madry2017towards}	&92.11 &65.05 &53.64 &52.18 &64.15 &6.6 &43.85 &0.51 &0.79 &55.12 &37.74 &14.51 &11.74 &10.83 &13.97 &19.43 &9.42 &0 &1.46 &11.8\\
			TRADES\cite{zhang2019theoretically}   &90.83 &\underline{66.27} &56.43 &55.15 &68.83 &6.25 &45.89 &1.96 &0.53 &\underline{57.46}	&34.8 &\textbf{16.8} &15.08 &14.7 &20.26 &20.7 &11.88 &0 &3.16 &\underline{15.25}\\
			MART\cite{wang2019improving} 	&92.01 &\textbf{69.02} &56.64 &54.78 &68.78 &12.04 &43.21 &1.87 &3.25 &\textbf{58.23} &36.56 &\underline{15.38} &13.33 &12.76 &16.52 &20.19 &10.63 &0 &2.28 &13.51\\
			\midrule
			TPAP(Ours) 	&\underline{94.09} &52.01 &\underline{90.73} &\underline{88.52} &\underline{94.73} &\underline{94.21} &\textbf{84.99} &\underline{80.45} &\underline{65.6} &47.65 &37.93 &11.69 &\underline{35.94} &\underline{35.41}       &35.71 &39.68 &\textbf{35.04} &\underline{33.69} &\textbf{19.46} &\textbf{23.03}\\
			TPAP+TRADES  &93.92 &57 &86.64 &79.05 &\textbf{95.5} &93.87 &73.23 &76.99 &53.35 &44.78 	&\textbf{50.43} &12.55 &27.56 &19.37 &\textbf{54.41} &\textbf{53.22} &25.01 &\textbf{43.88} &9.25 &6.93 \\
			TPAP+MART  &\textbf{94.24} &52.28 &\textbf{90.8} &\textbf{88.69} &94.51 &\textbf{94.23} &\underline{84.85} &\textbf{80.53} &\textbf{65.94} &48.35 &\underline{48.36} &8.46 &\textbf{41.56} &\textbf{35.69} &\underline{41.97} &\underline{44.3} &\underline{30.07} &16.34 &\underline{18.15} &14.66 \\	
			\bottomrule
		\end{tabular}
	}
	\vspace{3pt}
	\makebox[\textwidth]{\large} 
	\resizebox{\textwidth}{!}{
		\large
		\begin{tabular}{l|*{10}{c}|*{10}{c}}
			\toprule
			\fontsize{15}{20}\selectfont WideResNet-34 & \multicolumn{10}{c|}{\fontsize{15}{20}\selectfont CIFAR-10} & \multicolumn{10}{c}{\fontsize{15}{20}\selectfont CIFAR-100}\\ 
			\midrule
			Method & \Natural & \FGSM & \PGDT & \PGDO  & \CW & \DDN & \AAA & \STA & \FWA & \DIM & \Natural & \FGSM & \PGDT & \PGDO  & \CW & \DDN & \AAA & \STA & \FWA & \DIM \\
			\midrule
			PGD-AT\cite{madry2017towards} &\textbf{87.41} &59.27 &51.59 &50.7 &57.48 &19.28 &47.55 &0.11 &6.77 &52.24	&\underline{61.91} &32.75 &29.2 &28.63 &33.67 &10.16 &25.18 &0 &2.06 &29.53	\\
			TRADES\cite{zhang2019theoretically} &84.01 &\underline{60.04} &56.51 &56.13 &72.4 &21.78 &51.82 &0.62 &4.83 &\underline{56.86} 	&58 &33.26 &31.16 &30.81 &36.78 &12.42 &26.7 &0 &4.04 &31.17\\
			MART\cite{wang2019improving}	&\underline{85.67} &\textbf{61.98} &55.97 &55.15 &68.19 &21.37 &49.68 &0.88 &7.29 &\textbf{57.08}   &58.92 &\underline{35.14} &32.32 &31.92 &39.06 &12.62 &27 &0.01 &3.26 &\underline{32.73}\\
			\midrule
			TPAP(Ours) 	&82.73 &38.59 &\textbf{72.69} &\textbf{73.63} &58.28 &58.78 &\underline{67.23} &26.96 &19.97 &55.6 	&\textbf{64.34} &\textbf{38.49} &\underline{37.42} &\textbf{39.92} &42.87 &44.71 &\underline{38.97} &38.94 &\underline{20.72} &\textbf{33.37}\\
			TPAP+TRADES &84.13 &39.6 &60.03 &53.07 &\textbf{91.5} &\textbf{86.59} &60.25 &\textbf{79.52} &\underline{25.75} &28.59 &58.04 &31.88 &37.17 &33.38 &\textbf{73.17} &\textbf{66.71} &38.76 &\underline{65.67} &20.15 &20.48\\  	
			TPAP+MART &85.66 &36.27 &\underline{70.12} &\underline{64.29} &\underline{87.86} &\underline{81.63} &\textbf{70.55} &\underline{78.85} &\textbf{33.89} &23.47 	&56.36 &25.42 &\textbf{41.94} &\underline{38.77} &\underline{70.41} &\underline{62.54} &\textbf{42.28} &\textbf{67.58} &\textbf{25.89} &16.9\\
			\bottomrule
		\end{tabular}
	}
	\vspace{3pt}
	\makebox[\textwidth]{\large} 
	\resizebox{\textwidth}{!}{
		\large
		\begin{tabular}{l|*{10}{c}|*{10}{c}}
			\toprule
			\fontsize{15}{20}\selectfont WideResNet-34 & \multicolumn{10}{c|}{\fontsize{15}{20}\selectfont SVHN} & \multicolumn{10}{c}{\fontsize{15}{20}\selectfont Tiny-ImageNet}\\ 
			\midrule
			Method & \Natural & \FGSM & \PGDT & \PGDO  & \CW & \DDN & \AAA & \STA & \FWA & \DIM & \Natural & \FGSM & \PGDT & \PGDO  & \CW & \DDN & \AAA & \STA & \FWA & \DIM \\
			\midrule
			PGD-AT\cite{madry2017towards}	&92.88 &79.39 &71.2 &50.5 &66.12 &6.61 &37.21 &3.23 &0.73 &54.77    &53.3 &29.63 &26.84 &25.99 &33.7 &34.58 &21.81 &0.26 &1.43 &27.08\\
			TRADES\cite{zhang2019theoretically}  &\textbf{94.05} &\underline{83.61} &69.51 &60.7 &81.72 &5.95 &46.4 &1.66 &0.8 &\textbf{65.18}   &\underline{53.78}	&\underline{29.69}	&27.83	&27.1	&33.31	&34.25	&21.98	&0.25	&1.43	&\textbf{28.02}
			\\
			MART\cite{wang2019improving} &92.19 &82.85 &70 &51.78 &66.87 &9.04 &40.57 &6.21 &0.91 &\underline{58.02}     &52.18	&\textbf{29.95}	&27.8	&27.14	&34.17	&33.73	&21.86	&0	&1.54	&\underline{27.96}\\
			\midrule
			TPAP(Ours) 	&\underline{93.97} &\textbf{85.79} &\underline{84.85} &\textbf{87.16} &83.28 &75.92 &\textbf{89.12} &77.33 &\textbf{48.47} &57.86 	&48.04 &23.69 &\underline{30.46} &\textbf{29.25} &23.8 &35.32 &\underline{27.76} &4.49 &\textbf{36.32} &17.61\\
			TPAP+TRADES  &93.27 &45.68 &83.26 &73.37 &\textbf{93.83} &\textbf{89.19} &\underline{64.62} &\textbf{89.18} &4.55 &55.17   &\textbf{56.1} &17.31 &\textbf{33.17} &\underline{28.98} &\textbf{67.01} &\textbf{65.32} &\textbf{40.93} &\underline{61.49} &\underline{7.59} &10.94 \\
			TPAP+MART  &92.73 &21.45 &\textbf{88.07} &\underline{78.9} &\underline{88.47} &\underline{84.75} &62.77 &\underline{84.61} &\underline{7.54} &51.49 	&50.96 &10.29 &22.97 &21.79 &\underline{66.07} &\underline{62.08} &31.27 &\textbf{63.38} &{6.48} &5.19 \\	
			\bottomrule
		\end{tabular}
	}
\end{table*}
\section{Experiments}
\label{sec:Experiments}
We conduct comprehensive experiments on CIFAR-10, CIFAR-100 \cite{krizhevsky2009learning}, SVHN \cite{netzer2011reading}, Tiny-ImageNet \cite{le2015tiny} datasets with ResNet-18 \cite{he2016deep}, VGG-16 \cite{simonyan2014very} and WideResNet-34 \cite{zagoruyko2016wide} to evaluate the effectiveness of our proposed method.%, which include an empirical exploration, comparison experiments, ablation experiments and visualization experiments.
\subsection{Implementation Details}%
\label{sub:Implementation Details}
\subsubsection{Datasets}%
\label{sub:Datasets}
\begin{comment}
	\textbf{CIFAR-10} \cite{krizhevsky2009learning} consists of 10 common object classes, including 50,000 training images and 10,000 test images. The image size is 3×32×32 pixels.\par
	\textbf{CIFAR-100} \cite{krizhevsky2009learning} has 100 categories, with 600 color images per class, including 500 training images and 100 test images. The image size is 3×32×32 pixels.\par
	\textbf{SVHN} \cite{netzer2011reading}. The Street View House Numbers dataset is derived from house number plates in Google Street View images, with each image containing a sequence of Arabic numerals '0-9'. For the task of image classification, the images are resized and cropped to 3×32×32 pixels. It consists of 10 categories, with a total of 99,289 color images, including 73,257 training images and 26,032 testing images.\par
	\textbf{Tiny-ImageNet} \cite{le2015tiny} is a subset of the ImageNet dataset. It consists of 200 classes, with 500 training images, 50 validation images, and 50 test images per class. Notably, labels are not provided for the test examples. Each example in Tiny-ImageNet has a size of 3×64×64. \par
\end{comment}
%\textbf{CIFAR-10} \cite{krizhevsky2009learning} consists of 10 common object classes, including 50,000 training images and 10,000 test images. The image size is 3×32×32 pixels.\par
%\textbf{CIFAR-100} \cite{krizhevsky2009learning} has 100 categories, with 600 color images per class, including 500 training images and 100 test images. The image size is 3×32×32 pixels.\par
\textbf{CIFAR-10} and \textbf{CIFAR-100} \cite{krizhevsky2009learning} contain 10 and 100 categories, respectively.
\textbf{SVHN} \cite{netzer2011reading} is from house number plates in Google Street View images, containing a sequence of Arabic numerals `0-9'. 
%For the task of image classification, the images are resized and cropped to 3×32×32 pixels. It consists of 10 categories, with a total of 99,289 color images, including 73,257 training images and 26,032 testing images.\par
\textbf{Tiny-ImageNet} \cite{le2015tiny} is a subset of the ImageNet dataset. It consists of 200 classes.\par%, with 500 training images, 50 validation images, and 50 test images per class. Notably, labels are not provided for the test examples. Each example in Tiny-ImageNet has a size of 3×64×64. \par
%meaning it's a color image with a resolution of 64 pixels in both width and height. This dataset is commonly used for training and evaluating machine learning and computer vision models, especially in scenarios where the full ImageNet dataset may be too large or resource-intensive.\par
%For the experiments of CIFAR-10, CIFAR-100, SVHN and Tiny-ImageNet datasets, it’s necessary to normalize the pixel values to [0, 1] by dividing 255.
\subsubsection{Training Phase}%
\label{sub:Training Phase}
We set the batch size (bs) to 128 for CIFAR-10, CIFAR-100 and SVHN, and 64 for Tiny-ImageNet under ResNet-18. For VGG-16, we set the batch size as 128 for CIFAR-10, CIFAR-100, SVHN and Tiny-Imagenet. For WideResNet-34, we set the batch size to 64 for CIFAR-10, CIFAR-100 and SVHN, and 32 for Tiny-ImageNet.
We adopt stochastic gradient descent (SGD) \cite{robbins1951stochastic,ruder2016overview} optimizer with momentum factor of 0.9, an initial learning rate of 0.1 or 0.01 divided by 10 at the 75-th, 90-th and 140-th epochs and a weight decay factor of 1$\times$$10^{-3}$. The total number of epochs is set to 150. For CIFAR-10 and CIFAR-100, we augment the training data by random cropping and random horizontal flipping after filling 4 pixels. For Tiny-Imagenet, we only use random horizontal flipping.
We use FGSM-AT and set the maximum $\ell_\infty$-norm of adversarial perturbation to $\epsilon$ = 8/255 or 12/255. Adversarial purification radius is set to $\xi$ = 8/255. All experiments are implemented on the GeForce 1080 and 2080 TI GPU. % or $\ell_2$-norm 
\subsubsection{Evaluation Phase}%
\label{sub:Evaluation Phase}
%In order to verify the robust generalization ability of DNNs, 
We comprehensively consider various attack angles and potential vulnerabilities, and employ many different types of methods to generate adversarial examples to more comprehensively test and evaluate the overall robustness of deep learning models. 
We choose FGSM \cite{goodfellow2014explaining}, PGD \cite{madry2017towards} (PGD-20 and PGD-100), CW \cite{carlini2017towards}, DDN \cite{rony2019decoupling}, STA \cite{xiao2018spatially}, FWA \cite{wu2020stronger}, AutoAttack (AA) \cite{croce2020reliable} and TI-DIM \cite{dong2019evading,xie2019improving} attack methods in the white-box non-targeted attack setting. The code for attack methods comes from advertorch, torchattacks and authors. The adversarial perturbation strength of all attack methods under $\ell_\infty$-norm except CW and DDN under $\ell_2$-norm is set to 8/255.
We compare TPAP with the current mainstream adversarial training and image pre-processing methods \cite{yang2019me, hill2020stochastic, wang2023better}, including PGD-AT \cite{madry2017towards}, TRADES \cite{zhang2019theoretically}, MART \cite{wang2019improving} (Training the network with PGD-10) and SOAP \cite{shi2021online}. The combination of our proposed method and some of these methods are respectively referred to as TPAP+TRADES and TPAP+MART.

\subsection{Training Robust Overfitting Network}
\label{FGSM Robust Overfitting of DNNs}
\begin{figure}[t] \centering
	\begin{subfigure}{0.49\linewidth}
		\includegraphics[scale=0.1]{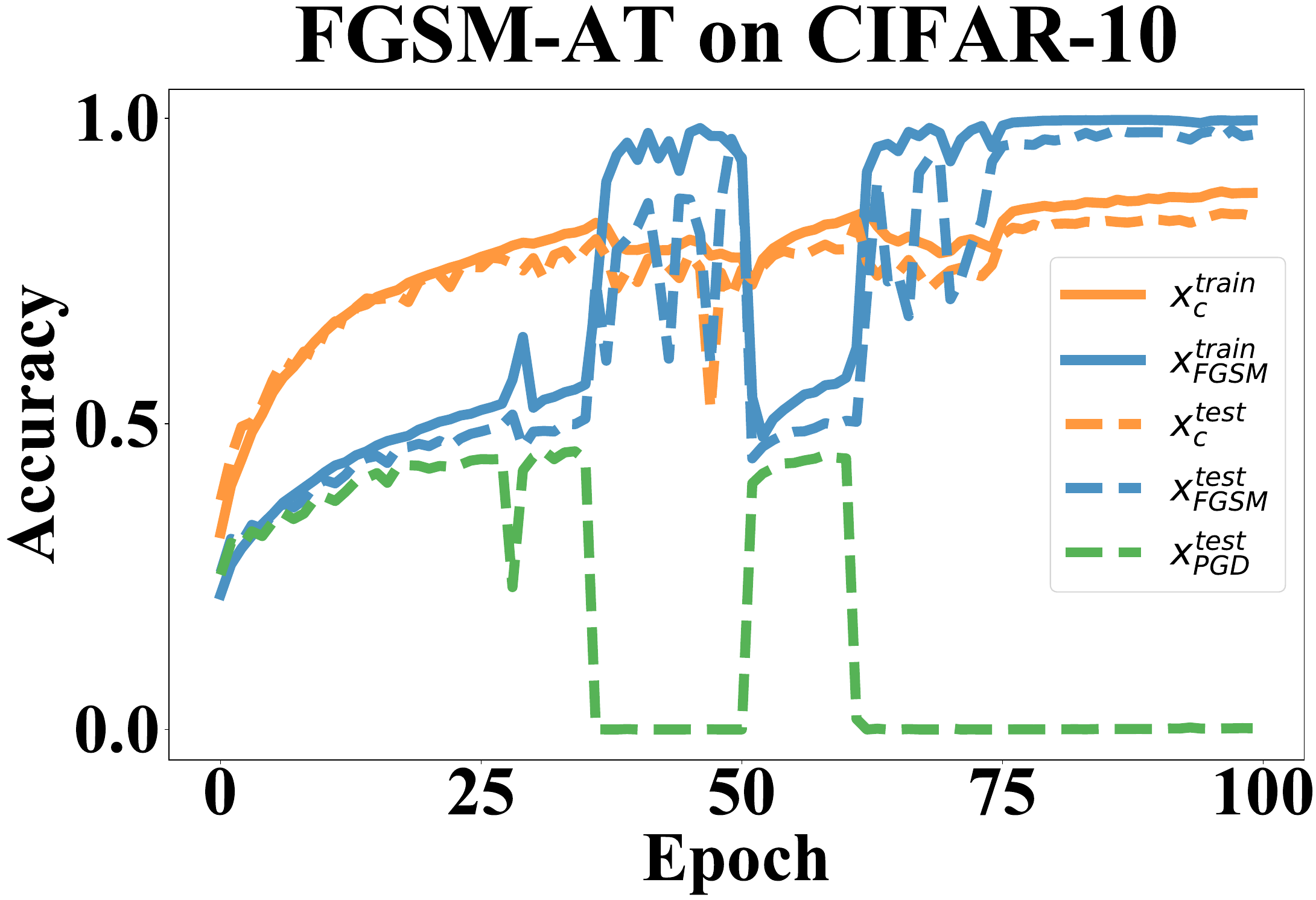}
		\caption{}
		\label{fig:figure3-a}
	\end{subfigure}
	\begin{subfigure}{0.49\linewidth}
		\includegraphics[scale=0.1]{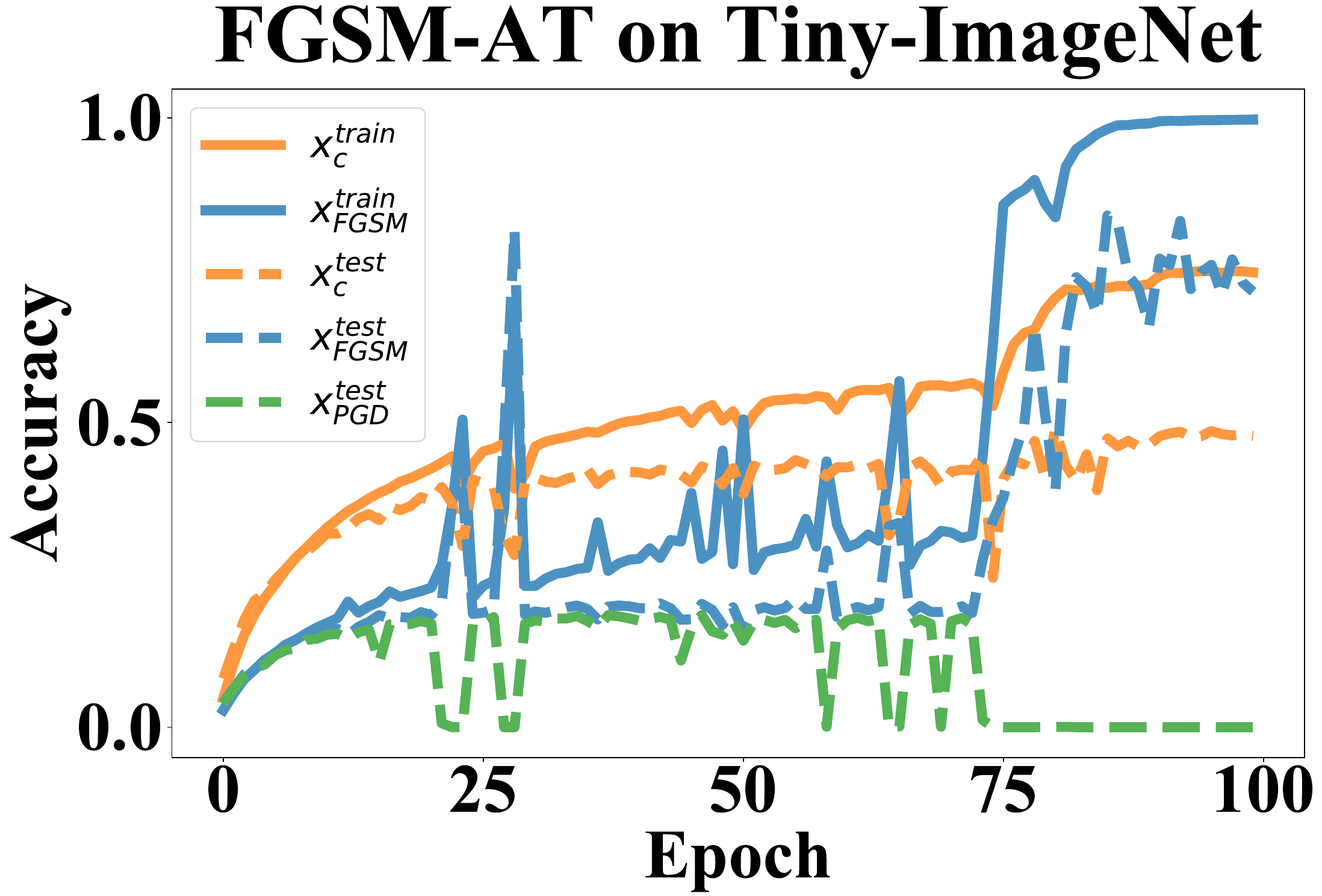}
		\caption{}
		\label{fig:figure3-b}
	\end{subfigure}
	\caption{(a) and (b) respectively represent the FGSM adversarial training process on the ResNet18 using CIFAR-10 and Tiny-ImageNet datasets. The horizontal axis represents epochs and the vertical axis represents classification accuracy. Solid lines represent the training set and dashed lines represent the test set.
		The orange, blue and green lines indicate the classification accuracy of the clean, FGSM and PGD adversarial examples, respectively.%, the FGSM adversarial examples, and the PGD adversarial examples, respectively.
	} 
	\label{fig:figure3}
\end{figure}

%Our method in the following section is based on FGSM robust overfitting prior of DNNs. However, e
Existing work has never explored to train a FGSM-RO-DNN to achieve robustness for both clean and adversarial examples. %optimal clean examples and FGSM adversarial examples correctness on the test set.
We find that three hyperparameters have a significant impact on FGSM robust overfitting, namely the learning rate of the network, the size of the training batch and the strength of the adversarial perturbation of the FGSM attack.
\cref{fig:figure3-a,fig:figure3-b} show FGSM adversarial training process, reflecting the effect of learning rate on the FGSM-RO-DNN. The initial learning rate is divided by 10 at the 75th and 90th epochs. A high learning rate leads to dramatic oscillations in the classification results of the adversarial examples, and as the learning rate decreases, the classification results of the clean and FGSM adversarial examples gradually increase and stabilize.
\begin{figure}[t] \centering
	\begin{subfigure}{0.49\linewidth}
		\includegraphics[scale=0.13]{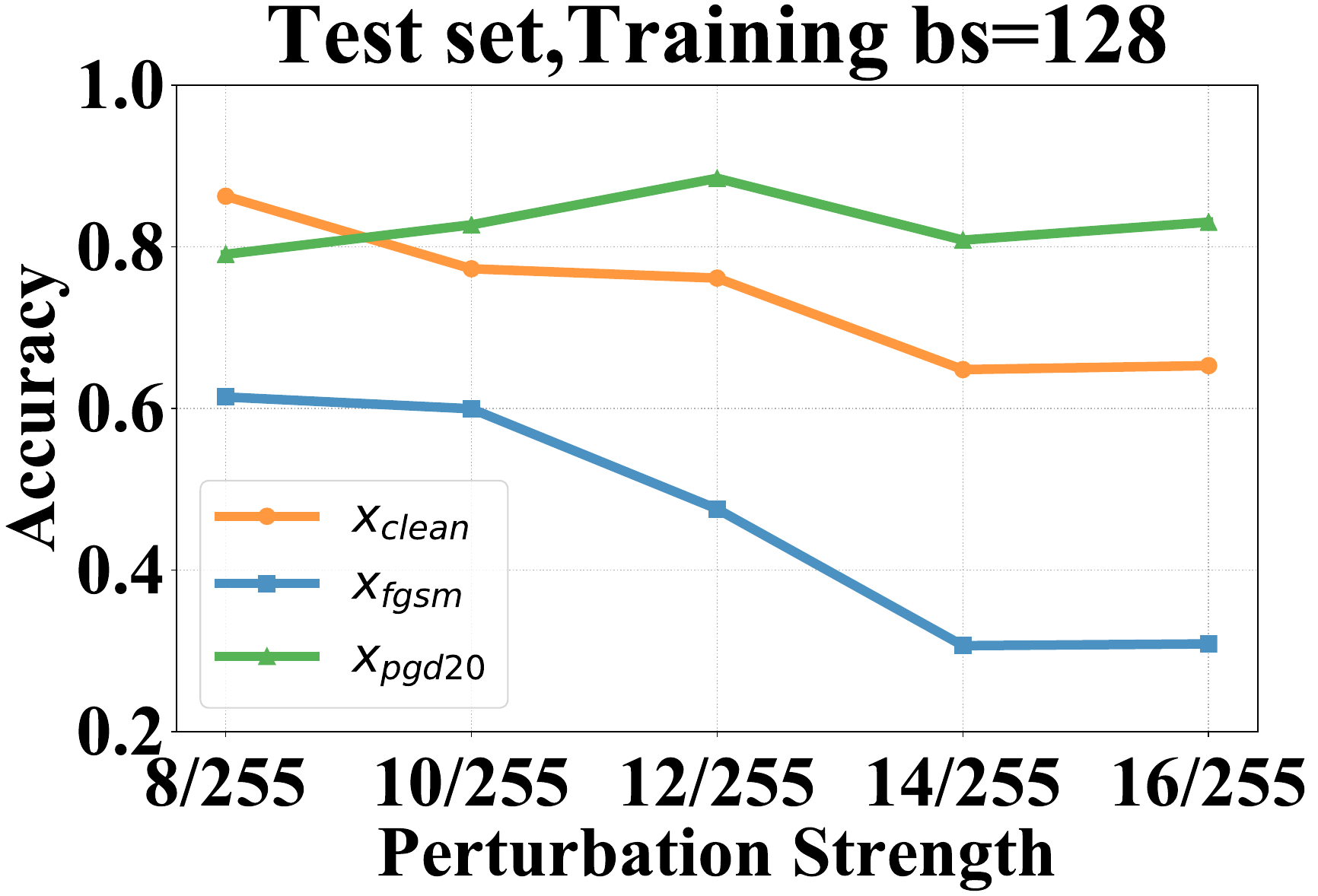}
		\caption{}
		\label{fig:figure6-a}
	\end{subfigure}
	\begin{subfigure}{0.49\linewidth}
		\includegraphics[scale=0.13]{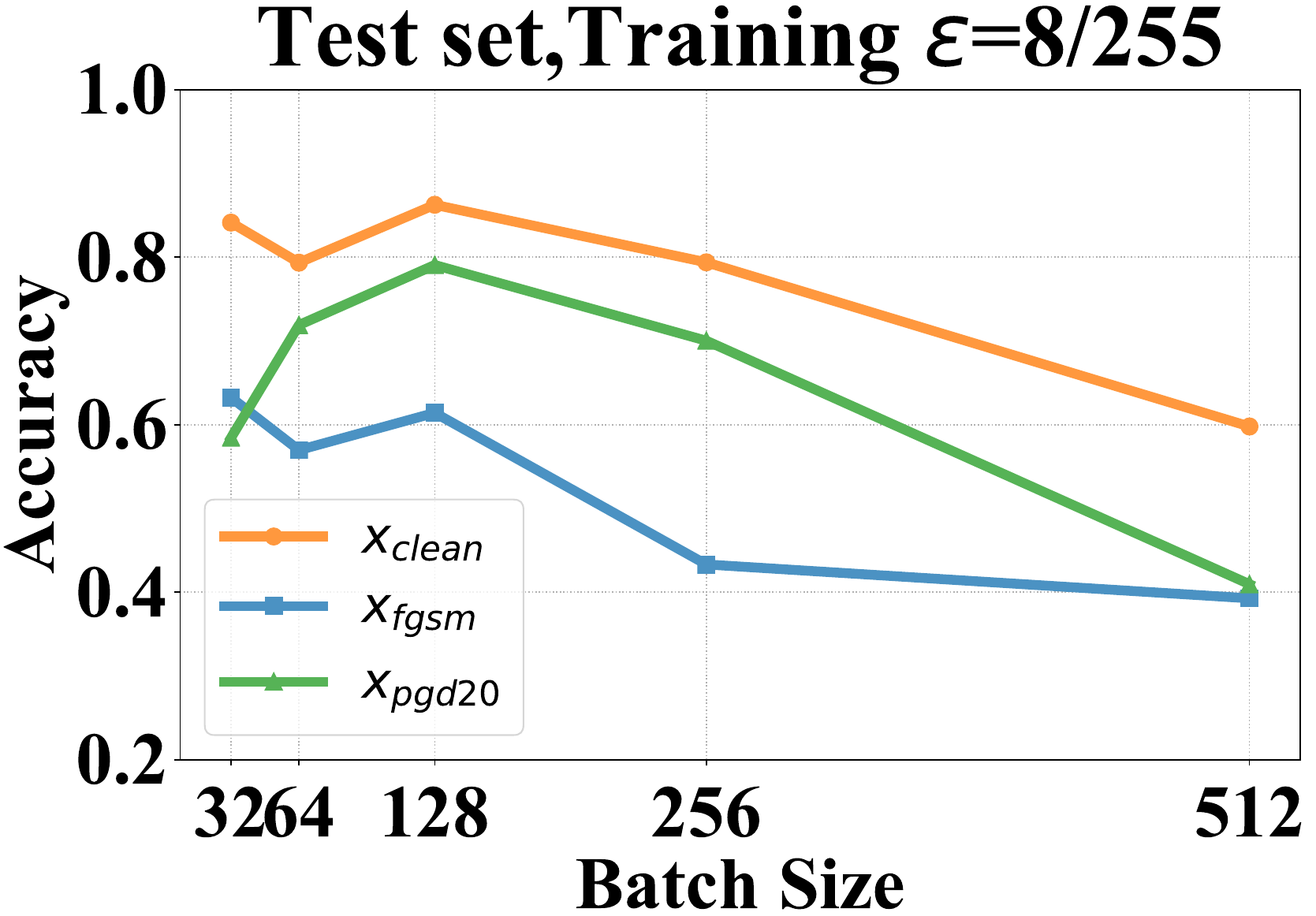}
		\caption{}
		\label{fig:figure6-b}
	\end{subfigure}
	\\
	\caption{(a) and (b) represent ablation study of robust training in TPAP under various perturbation strengths and batch sizes.
	} 
	\label{fig:figure6}
\end{figure}

%Within a reasonable range, the larger the training batch size is, the more accurate the direction of gradient descent it determines, and the less training oscillations it causes, so that the FGSM robust overfitting network has better robustness and generalization ability.
%Exploratory experiments show that sometimes the network works well on the training set but has low classification correctness on the test set because the network is overfitting to the FGSM adversarial examples on the training set, and increasing the perturbation strength of the FGSM adversarial examples on the training set (from 8/255 to 12/255) dramatically enhances the network's generalization to the FGSM adversarial examples on the test set.
% 受参数影响的消融实验
\cref{fig:figure6-a,fig:figure6-b} explore the effects of perturbation strength and batch size on robust overfitting characteristic of DNN, respectively.
We conduct an ablation study on CIFAR-10 dataset using ResNet-18 to explore these key hyperparameters. The figures respectively illustrate line plots depicting the variation in classification results during the training phase as the perturbation strength ranges from 8/255 to 16/255 with a step size of 2/255 and the batch size varies across 32, 64, 128, 256 and 512. The ablation results show that a trade-off between $\epsilon$ and batch size is required to obtain FGSM-RO-DNN in TPAP. 

\subsection{Experimental Results}%
\label{sub:Experimental results on DNNs}
%We propose for the first time a defense method for training FGSM robust overfitting neural networks and purifying input images during the test phase. 
The white-box attack results for CIFAR-10, CIFAR-100, SVHN and Tiny-ImageNet are shown in \cref{tab:table1}.
%From the experimental results, we can see that our proposed method outperforms other methods in most attack scenarios, not only improving the accuracy of the network on clean data, but also further improving the accuracy on the various attacks of adversarial examples. 
On these datasets, TPAP and its variants maintain higher accuracy on clean examples compared to AT based methods.
%, especially reaching 86.18\% on cifar10 and 61.03\% on cifar100. 
In terms of adversarial robustness, PGD adversarial training network cannot resist STA and FWA attacks, but TPAP greatly improves the accuracy of adversarial examples under these attacks. %For the attack methods used in the experiments, 
Also, TPAP outperforms other methods in most adversarial attack scenarios.
This is because most existing defense methods ignore the diversity of attacks. Our method can perform pixel-level purification of adversarial examples generated by unknown attacks in a ``counter changes with changelessness" manner, and reliably pulls the adversarial examples back within the boundary of correct classification. This benefits from the vulnerability of the FGSM-RO-DNN to non-FGSM adversarial examples and the robustness to FGSM adversarial examples.

However, TPAP is usually less robust to 8/255 FGSM adversarial examples than AT, which is explainable and easy to understand as shown in $x_{FGSM}$ of \cref{fig:figure7-a}. The image purification process makes the correctly classified 8/255 FGSM examples subject to FGSM attack far from the correct label, which is amount to generating 16/255 FGSM adversarial examples. But the network is not trained on 16/255 FGSM adversarial examples, so it cannot classify them with high accuracy. 
To deal with this problem, we try to use both 8/255 and 16/255 FGSM adversarial examples to train the network, and the results are shown in \cref{tab:table2}. 
Although it enhances the robustness of 8/255 FGSM adversarial examples, it reduces the classification accuracy on clean examples from 86.25\% to 82.11\%.
Further, we use clean examples, 8/255 and 16/255 FGSM adversarial examples to train the network, but may not obtain a FGSM-RO-DNN.

Experiments further validate the robust overfitting of TPAP acting on large-size images, such as Caltech-101, consisting of a total of 9146 images from 101 object classes, as well as an additional background/clutter class. The image size is 300$\times$200. Each object category contains between 40 and 800 images on average.
To account for domain gap within Caltech-101, we utilize the ResNet-18 network pre-trained on ImageNet provided by official PyTorch implementation. In \cref{tab:table4}, * indicates the use of pre-trained network. The train and test sets are split randomly in 8/2 ratio. Compared with PGD-AT* ($\epsilon$=8/255), TPAP-TRADES* ($\epsilon$=16/255, bs=32) shows superior robustness on both clean examples and PGD-adversarial examples ($\epsilon$=8/255).
More comparative experiments with image pre-processing methods on CIFAR-10 are presented in \cref{tab:table5}. 
\vspace{-5pt}
\begin{table}[htbp]\centering \large
	\caption{Classification accuracy against adversarial examples on CIFAR-10. TPAP denotes the robust overfitting ResNet-18  trained with 8/255 FGSM adversarial examples, while TPAP* includes both 8/255 and 16/255 FGSM adversarial examples.}
	\label{tab:table2}
	\resizebox{0.48\textwidth}{!}{
		\large
		\begin{tabular}{l|*{10}{c}}
			\toprule
			ResNet-18 & \multicolumn{10}{c}{CIFAR-10} \\ 
			\midrule
			Method & Clean & FGSM & PGD-20 & PGD-100  & CW$_2$ & DDN & AA & STA & FWA & TI-DIM  \\
			\midrule
			TPAP  &{86.25}	&{61.41}	&{79.06}	&{80.5}	&61.37	&64.5	&{76.34}	&31.4	&{52.83}	&{75.21}  \\
			TPAP*  &82.11	&73.23	&80.84	&79.69	&80.65	&79.04	&75.72	&69.87	&66.08	&80.85	\\
			%FGSM-AT*	&62.07		&44.58		&45.54		&48.81		&70.75		&84.54		&76.92		&75.39		&10.9	&44.07 \\
			\bottomrule
		\end{tabular}
	}
\end{table}
\vspace{-12pt}
\begin{center}
	\begin{minipage}[b]{0.42\linewidth}
		\centering
		\captionof{table}{ Caltech-101.}
		\label{tab:table4}
		\resizebox{\textwidth}{!}{
			\begin{tabular}{l|*{3}{c}}
				%\toprule
				%ResNet-18 & \multicolumn{3}{c}{CALTECH-101($\epsilon$=8/255)} \\ 
				\midrule
				Method & Clean    &FGSM    &   PGD-20  \\
				\midrule
				PGD-AT* & 72.44& {65.55} &57.74              \\
				TPAP* & 70.2& 55.93& 59.16                \\
				TPAP-TRADES* & {76.11}& 60.8& {70.91}           \\
				TPAP-MART* & 69.27& 58.39& 60.63          \\
				\bottomrule
			\end{tabular}
		}
	\end{minipage}\hfill
	\begin{minipage}[b]{0.55\linewidth}
		\centering
		\captionof{table}{ Comparative experiment.}
		\label{tab:table5}
		\resizebox{\textwidth}{!}{
			\begin{tabular}{l|*{3}{c}}    %*{1}{c}
				\toprule
				Method & Clean & PGD & Architecture \\ 
				\midrule
				(Yang et al., 2019)(p:0.4$\to$0.6) \cite{yang2019me} & 84  & 68.2 & ResNet-18           \\ 
				(Hill et al., 2021) \cite{hill2020stochastic} &  84.12 &78.91    & WRN-28-10  \\
				(Wang et al., 2023) \cite{wang2023better} & 92.58 & 68.43 & WRN-28-10            \\      
				\midrule
				TPAP & 86.25 & 79.06 & ResNet-18               \\
				\bottomrule
			\end{tabular}
		}
	\end{minipage}
\end{center}	
%Experiments on VGG-16 and WideResNet-34 are presented in Appendix. %More experimental details are included in the supplementary material.
\vspace{1pt}
\subsection{Computational overhead and ablation study}%
% 计算开销和训练时间
We use DeepSpeed from Microsoft to compute FLOPs for TPAP and PGD-AT (trained on PGD-10 adversarial examples) in the same conditions, presented in \cref{tab:table3}. Compared with PGD-AT, TPAP reduces computation cost on adversarial examples generation during training, but increases in testing phase due to the adversarial purification operation. This is common because the proposal is a test-time approach.
\cref{tab:table6} presents ablation experiment of TPAP and verifies the RO-FGSM-DNN performs well on both clean data and FGSM adversarial examples. After adversarial purification, the accuracy of TPAP for clean data just reduces by 0.08\%. This indicates our model is plastic to clean data.
Furthermore, we replace FGSM attack with CW$_2$ and PGD attacks, and as shown in \cref{tab:table7}, the accuracy on clean samples is decreased. This further supports our finding that FGSM attack is unique for robust overfitting.
\vspace{-6pt}
\begin{table}[htbp]\centering \large
	\caption{Comparison of computational overhead.}
	\label{tab:table3}
	\resizebox{0.48\textwidth}{!}{
		\large
		\begin{tabular}{l|*{5}{c}}
			\toprule
			ResNet-18 & \multicolumn{5}{c}{CIFAR-10($\epsilon$=8/255, Batch size = 128)} \\ 
			\midrule
			Method & Training time (s)/epoch & Test time (s)/epoch & Training FLOPs (T)/epoch & Test FLOPs (T)/epoch  & Params (M) \\
			\midrule
			PGD-AT  &{273.89}	&{2.02}	&{1831.5}	&{11.1}	&11.17		  \\
			TPAP  &66.26	&8.26	&333	&44.4	&11.17			\\
			\bottomrule
		\end{tabular}
	}
\end{table}
\vspace{-16pt}
\begin{center}
	\begin{minipage}[t]{0.53\linewidth}
		\centering
		\captionof{table}{\scriptsize Ablation study of TPAP.}
		\label{tab:table6}
		\resizebox{\textwidth}{!}{
			\begin{tabular}{*{2}{c}|*{3}{c}}
				\toprule
				\multicolumn{2}{c}{ResNet-18} & \multicolumn{3}{|c}{CIFAR-10($\epsilon$=8/255,bs=128)} \\ 
				\midrule
				Purification & RO-FGSM-DNN	&Clean &FGSM	&	PGD-20  \\
				\midrule
				$\times$ & \checkmark & 86.33	&94.41	&0.18			  \\
				\checkmark & \checkmark &86.25	&61.41	&79.06				\\
				\bottomrule
			\end{tabular}
		}
	\end{minipage}\hfill
	\begin{minipage}[t]{0.44\linewidth}
		\centering
		\captionof{table}{\scriptsize Under different attacks.}
		\label{tab:table7}
		\resizebox{\textwidth}{!}{
			\begin{tabular}{l|*{5}{c}}
				\toprule
				ResNet-18 & \multicolumn{3}{|c}{CIFAR-10($\epsilon$=8/255,bs=128)} \\ 
				\midrule
				Training & Clean &CW2	&PGD-20 &FGSM	&AA  \\
				\midrule
				CW2 & 68.19 & 84.7 & 37.4 & 39.95 & 56.66			  \\
				PGD-10 & 56.17 & 75.62 & 42.74 & 44.5 & 82				\\
				FGSM & 86.25 & 61.37 & 79.06 & 61.41 & 76.34 				\\
				\bottomrule
			\end{tabular}
		}
	\end{minipage}
\end{center}
\subsection{Analysis of the Visualization Experiments}
\label{sub:Analysis of the Visualization Experiments}
\begin{figure}[t] \centering
	%\small
	\includegraphics[width=0.10\textwidth]{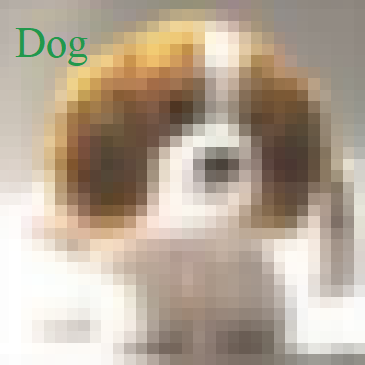}
	\includegraphics[width=0.10\textwidth]{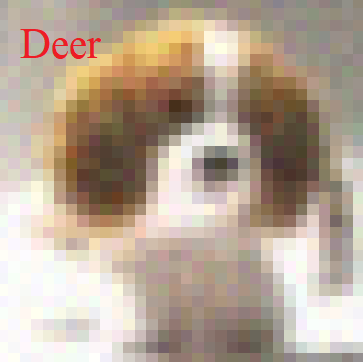}
	\includegraphics[width=0.10\textwidth]{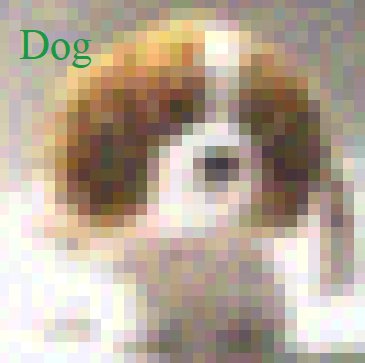}
	\includegraphics[width=0.10\textwidth]{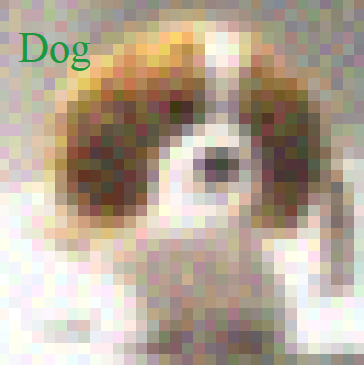}
	\\
	\vspace{1pt}
	\includegraphics[width=0.10\textwidth]{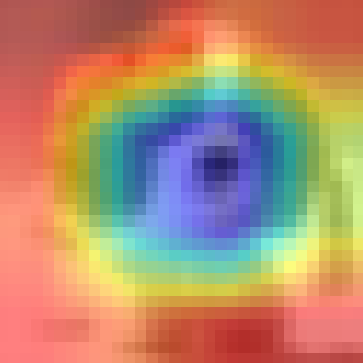}
	\includegraphics[width=0.10\textwidth]{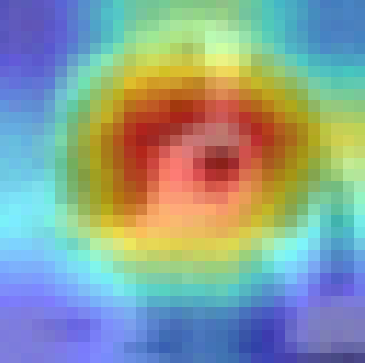}
	\includegraphics[width=0.10\textwidth]{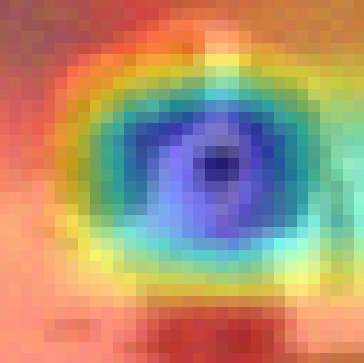}
	\includegraphics[width=0.10\textwidth]{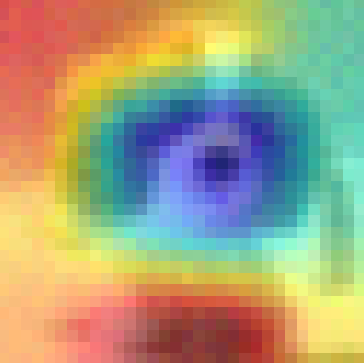}
	\\
	\vspace{1pt}
	\includegraphics[width=0.10\textwidth]{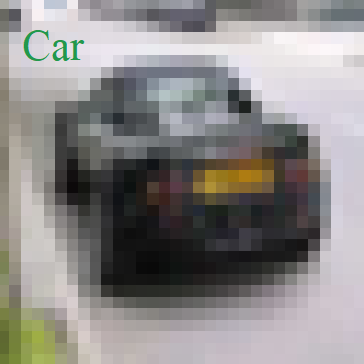}
	\includegraphics[width=0.10\textwidth]{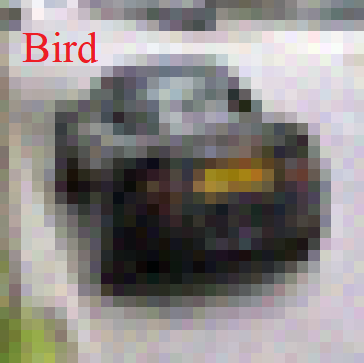}
	\includegraphics[width=0.10\textwidth]{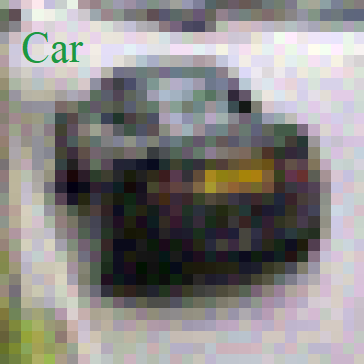}
	\includegraphics[width=0.10\textwidth]{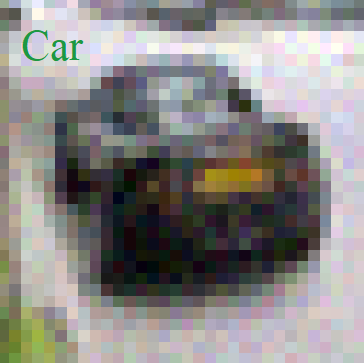}
	\\
	\vspace{1pt}
	\includegraphics[width=0.10\textwidth]{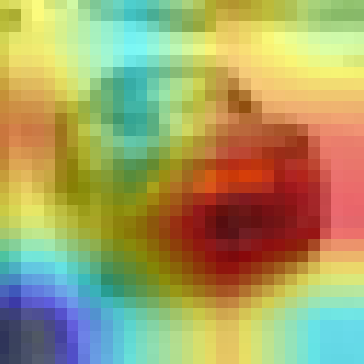}
	\includegraphics[width=0.10\textwidth]{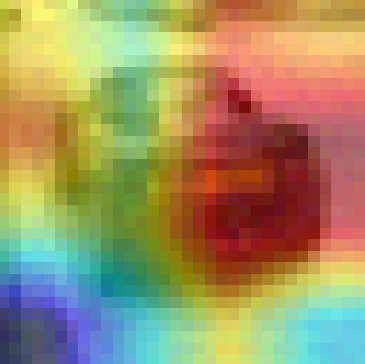}
	\includegraphics[width=0.10\textwidth]{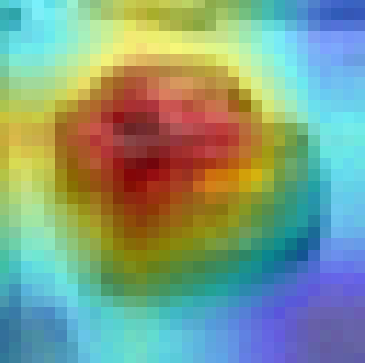}
	\includegraphics[width=0.10\textwidth]{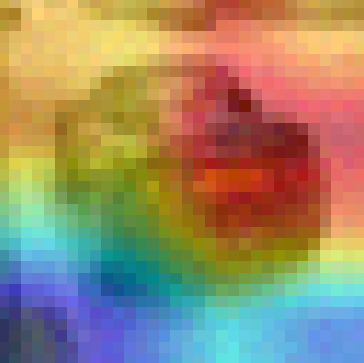}
	\\
	\vspace{1pt}
	\makebox[0.10\textwidth]{\scriptsize (a) $x_{c}$}
	\makebox[0.10\textwidth]{\scriptsize (b) $x_{pgd20}$}
	\makebox[0.10\textwidth]{\scriptsize (c) $x_{c\_pur}$}
	\makebox[0.10\textwidth]{\scriptsize (d) $x_{pgd20\_pur}$}
	\\
	\caption{Attention visualization of TPAP. We show 2 examples of \textit{dog} and \textit{car} respectively.
		The first and second column respectively indicate the attention map of clean and the PGD-20 adversarial examples, represented as $x_{c}$ and $x_{pgd20}$. %, don't pass through purification but are instead directly fed into the FGSM-RO-DNN, along with their respective attention areas.
		The third and fourth column respectively represents the attention maps of purified examples $x_{c\_pur}$ and $x_{pgd20\_pur}$ obtained by our TPAP.}
	\label{fig:figure4}
\end{figure}

\textbf{Grad-CAM} \cite{selvaraju2017grad} \textbf{Visualization}. Grad-CAM
%, short for Gradient-weighted Class Activation Mapping, 
is a visualization technique used to explain the decision-making process of models, highlighting crucial regions of an image feature map. %for the model's classification decisions. 
\cref{fig:figure4} demonstrates the visualization experiment of TPAP. %Columns (a)(b) do not have the test-time purification process but columns (c)(d) do.
We see the adversarial examples in (b) have biased attention regions compared to those of correctly classified examples in (a), (c) and (d), although their attention regions can focus on the target object.
The results in (a), (c) and (d) show that the FGSM-RO-DNN focuses on the global outline or internal regions of the target category for correctly classified examples.
%\cref{fig:figure4}(c)(d) show that both clean examples and adversarial examples, under TPAP purification, can be correctly classified, with the attention span encompassing the edges and specific internal regions of the target category. 
There are two puzzling questions: \textit{Why does the FGSM-RO-DNN focus on the global outline of the target category?} and \textit{why does the FGSM-RO-DNN result in misclassification when focusing on the interior of the target object?} In low-resolution image classification tasks, there is minimal spurious correlation between labels and backgrounds due to the similarity in background colors for different classes. The network finds it challenging to learn correct classification knowledge from non-causal factors such as background. Therefore, we argue that the FGSM-RO-DNN learns one of the most significant distinctions between categories, i.e. the global outline. 
Additionally, different target categories share similar internal regions, such as cats and dogs having similar fur colors and facial features, or cars and trucks having similar colors. Meanwhile, the FGSM-RO-DNN is highly vulnerable to other types of attacks except FGSM, which also leads to misclassification even focused on interior of target class.
\begin{figure}[t] \centering
	PGD-AT
	\\
	\vspace{3pt}
	\includegraphics[width=0.105\textwidth]{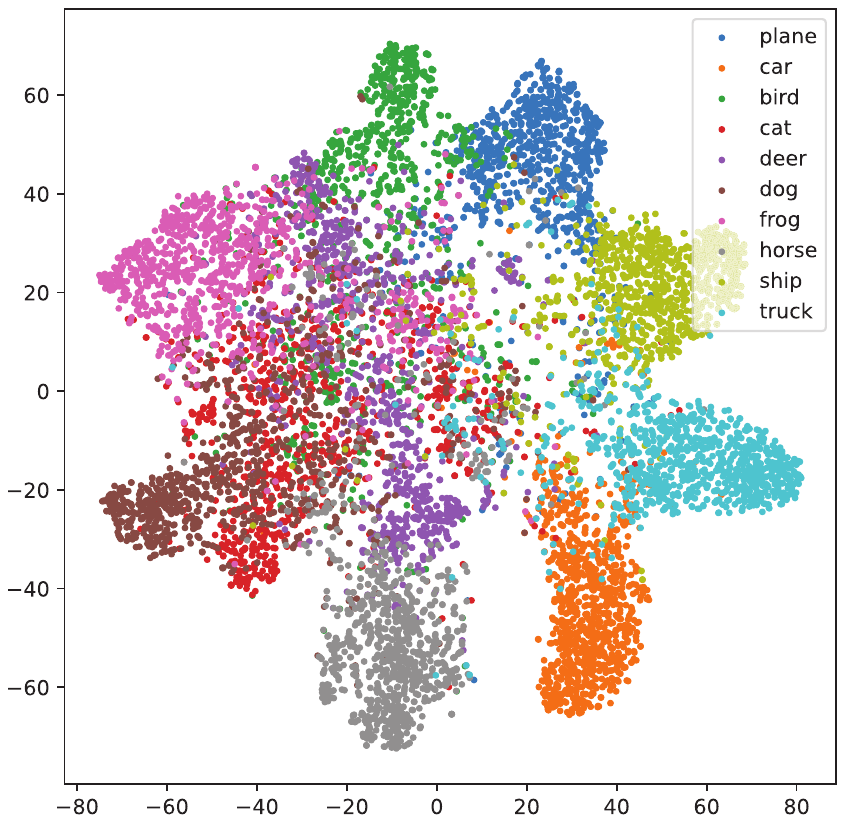}
	\includegraphics[width=0.105\textwidth]{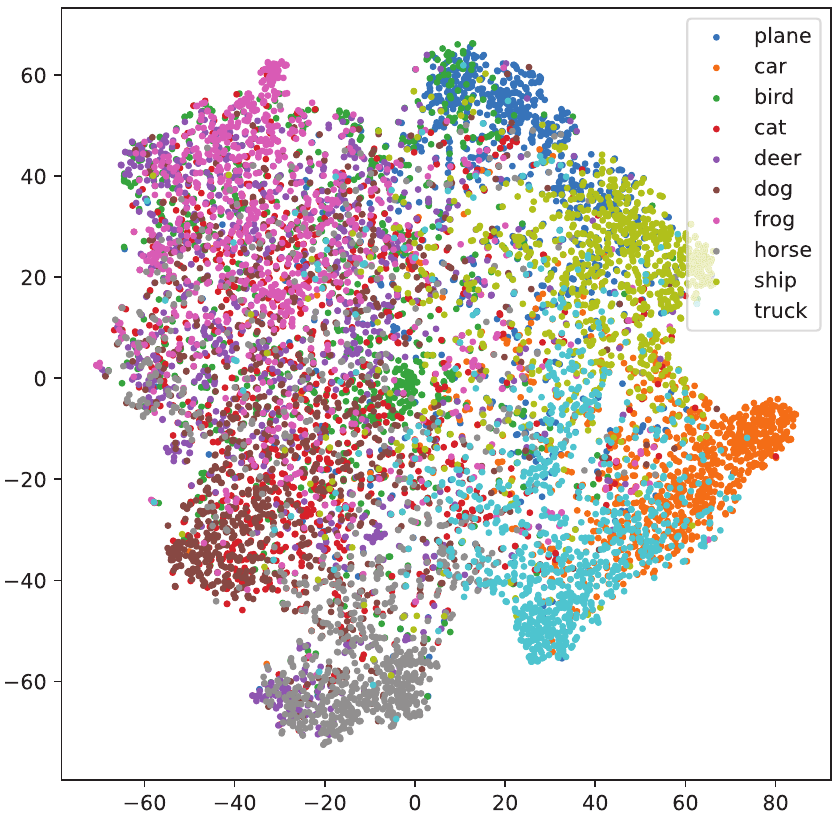}
	\includegraphics[width=0.105\textwidth]{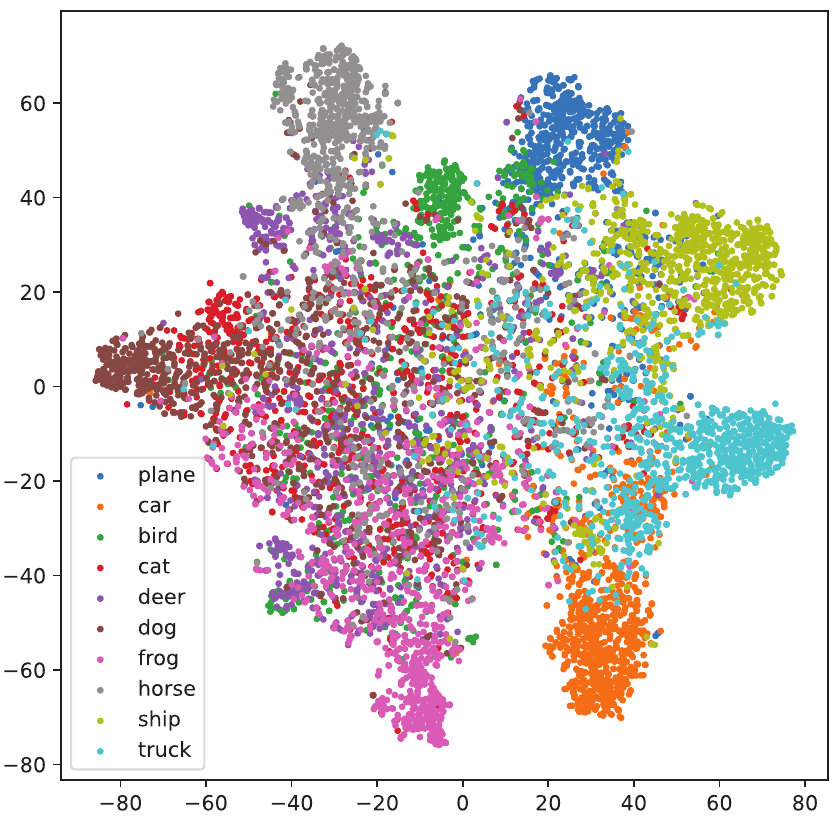}
	\includegraphics[width=0.105\textwidth]{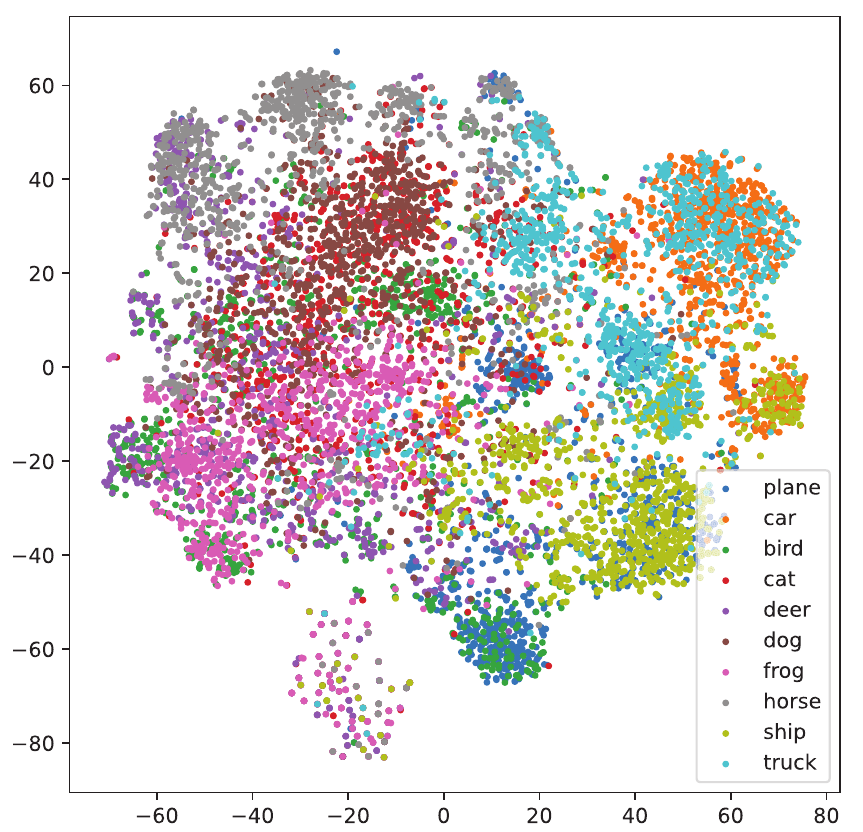}
	\vspace{6pt}
	\\
	TPAP
	\\
	\vspace{3pt}
	\includegraphics[width=0.105\textwidth]{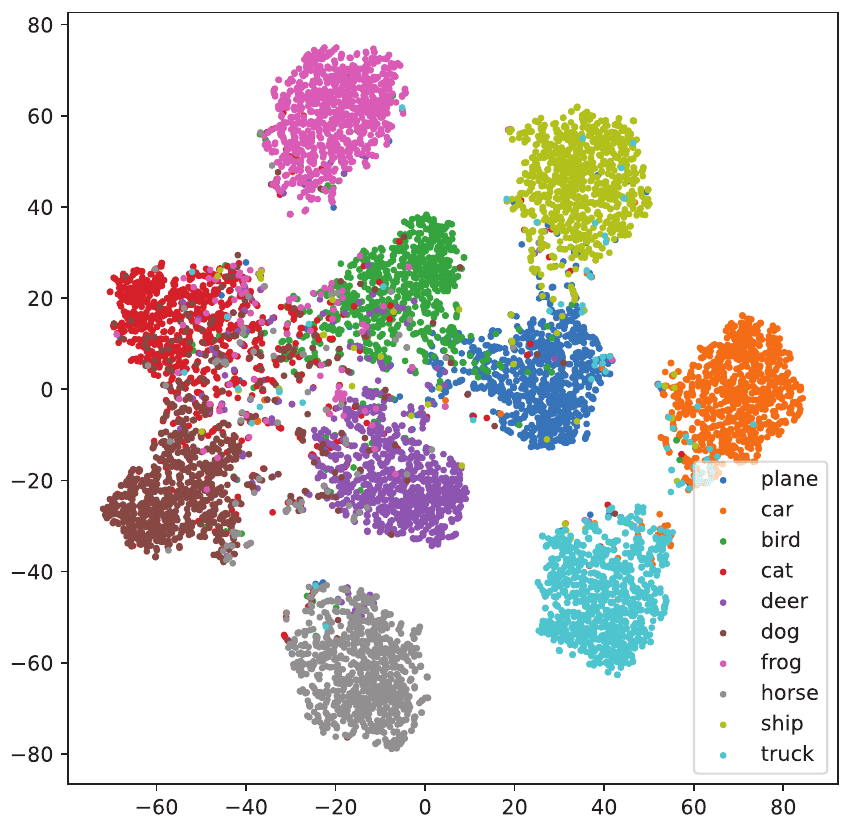}
	\includegraphics[width=0.105\textwidth]{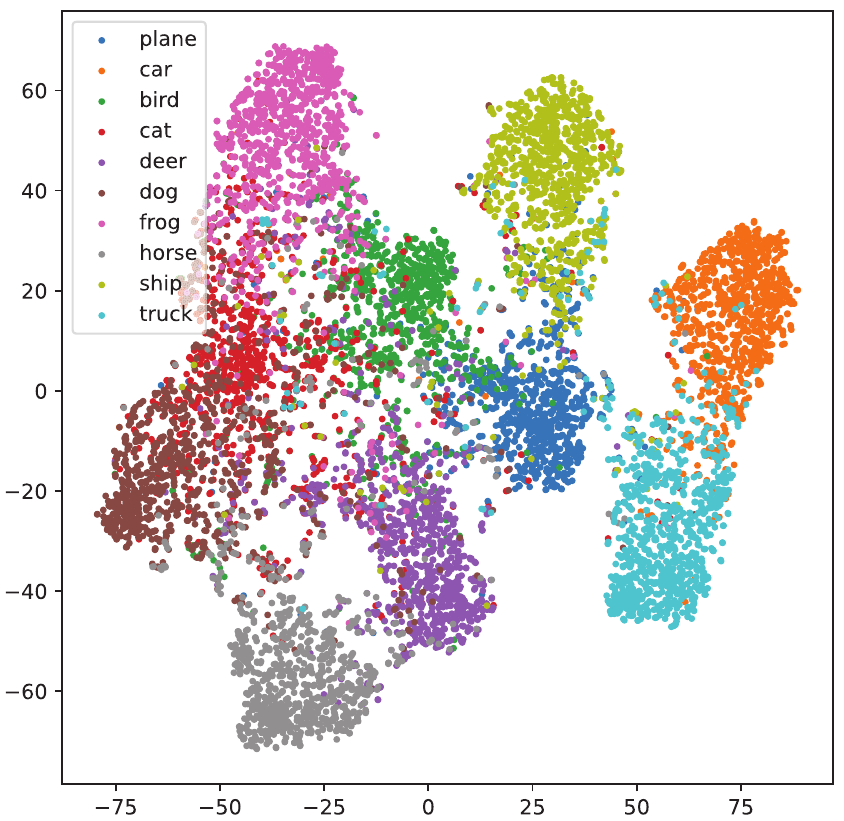}
	\includegraphics[width=0.105\textwidth]{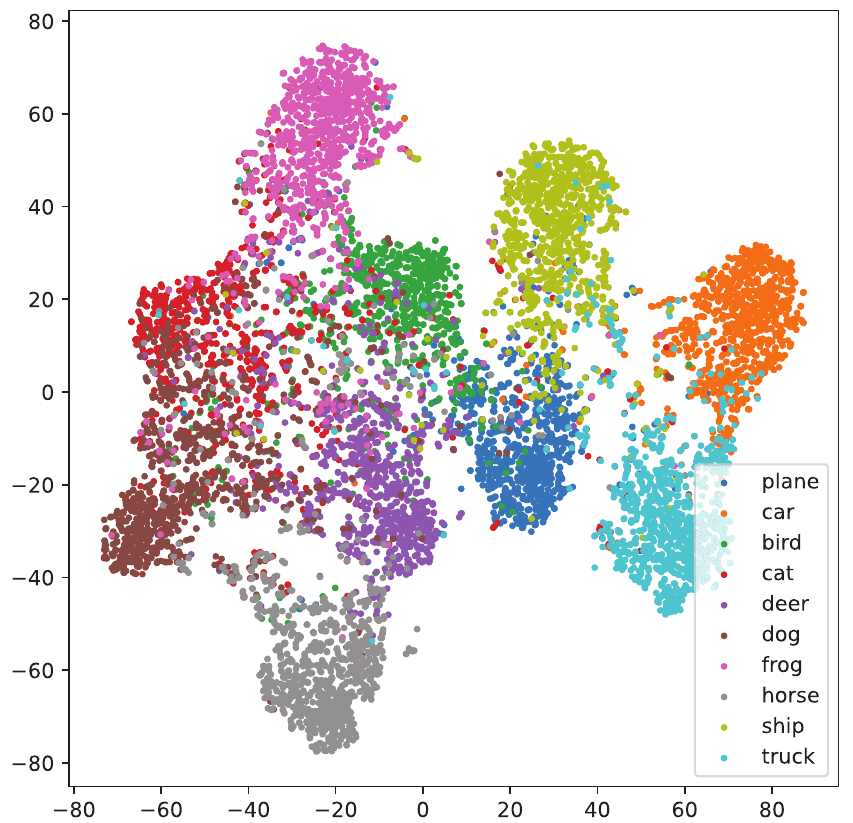}
	\includegraphics[width=0.105\textwidth]{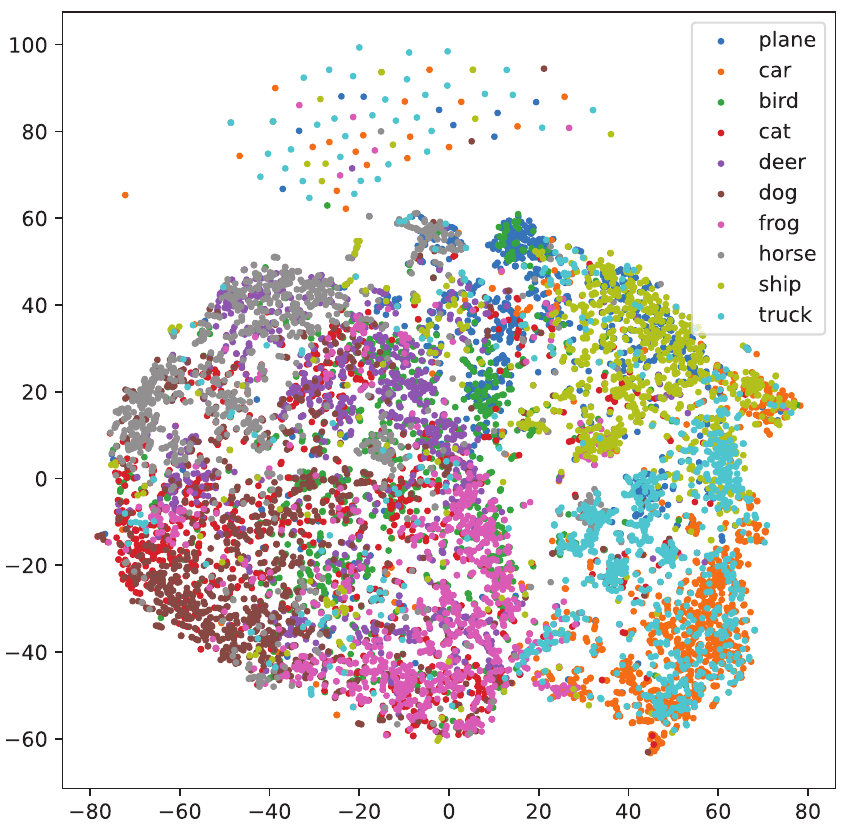}
	\\
	\makebox[0.10\textwidth]{\scriptsize (a) $x_{c}$}
	\makebox[0.10\textwidth]{\scriptsize (b) $x_{pgd20}$}
	\makebox[0.10\textwidth]{\scriptsize (c) $x_{AA}$}
	\makebox[0.10\textwidth]{\scriptsize (d) $x_{STA}$}
	\\
	\caption{t-SNE feature distribution visualization of PGD-AT and our proposed TPAP on clean and adversarial examples.} 
	\label{fig:figure5}
\end{figure}
\\ \indent\textbf{Feature Visualization}. \cref{fig:figure5} visualizes the feature distribution from the penultimate layer of network. We use t-SNE \cite{van2008visualizing} to project CIFAR-10 features onto a two-dimensional plane, where the top row comes from the baseline (PGD-AT) and the second row represents TPAP. We observe adversarial attacks often distort the discriminative feature distribution of PGD-AT network, whereas TPAP can effectively improve the adversarial distributions with better feature clustering (i.e., inter-class separability and intra-class compactness). Robustness is indicated.% consistent with the semantic cognition of the target network.
\section{Conclusion}%
\label{sec:Conclusion}
We propose a novel adversarial defense method and, for the first time, introduce FGSM robust overfitting to instruct test-time robustness. TPAP is easy to train and computationally efficient, and remarkably enhances the robust generalization. We hope our work can inspire future research.\par
\textbf{Limitations}. TPAP has two limitations. Firstly, TPAP does not perform well in defending against 8/255 FGSM adversarial examples, and the reason is explained in Section \ref{sub:Experimental results on DNNs}.
Secondly, TPAP fails to defend against attacks of too strong perturbations, such as $\epsilon$=0.3 on MNIST \cite{lecun1998gradient} dataset. 
%Because essentially, TPAP is based on the attack, called adversarial purification.
This is due to existing adversarial attacks move clean examples away from correctly classified labels, while our adversarial purification moves adversarial examples away from misclassified labels.
When the adversarial perturbation is large, the force of adversarial purification also needs to be enlarged, leading to large pixel changes in image. This severely destroys the semantics of the image and leads to a significant decrease in classification accuracy. \\
\indent \textbf{Acknowledgement.} 
This work was supported by National Key R\&D Program of China (2021YFB3100800), National Natural Science Fund of China (62271090), Chongqing Natural Science Fund (cstc2021jcyj-jqX0023) and National Youth Talent Project. %This work is also supported by Huawei computational power of Chongqing Artificial Intelligence Innovation Center.
\clearpage
%%%%%%%%% REFERENCES
{\small
	\bibliographystyle{ieee_fullname}
	\bibliography{main}

\begin{thebibliography}{10}\itemsep=-1pt

\bibitem{akhtar2018defense}
Naveed Akhtar, Jian Liu, and Ajmal Mian.
\newblock Defense against universal adversarial perturbations.
\newblock In {\em Proceedings of the IEEE conference on computer vision and
  pattern recognition}, pages 3389--3398, 2018.

\bibitem{andriushchenko2020understanding}
Maksym Andriushchenko and Nicolas Flammarion.
\newblock Understanding and improving fast adversarial training.
\newblock {\em Advances in Neural Information Processing Systems},
  33:16048--16059, 2020.

\bibitem{carlini2017towards}
Nicholas Carlini and David Wagner.
\newblock Towards evaluating the robustness of neural networks.
\newblock In {\em 2017 ieee symposium on security and privacy (sp)}, pages
  39--57. Ieee, 2017.

\bibitem{croce2020reliable}
Francesco Croce and Matthias Hein.
\newblock Reliable evaluation of adversarial robustness with an ensemble of
  diverse parameter-free attacks.
\newblock In {\em International conference on machine learning}, pages
  2206--2216. PMLR, 2020.

\bibitem{dong2018boosting}
Yinpeng Dong, Fangzhou Liao, Tianyu Pang, Hang Su, Jun Zhu, Xiaolin Hu, and
  Jianguo Li.
\newblock Boosting adversarial attacks with momentum.
\newblock In {\em Proceedings of the IEEE conference on computer vision and
  pattern recognition}, pages 9185--9193, 2018.

\bibitem{dong2019evading}
Yinpeng Dong, Tianyu Pang, Hang Su, and Jun Zhu.
\newblock Evading defenses to transferable adversarial examples by
  translation-invariant attacks.
\newblock In {\em Proceedings of the IEEE/CVF Conference on Computer Vision and
  Pattern Recognition}, pages 4312--4321, 2019.

\bibitem{eykholt2018robust}
Kevin Eykholt, Ivan Evtimov, Earlence Fernandes, Bo Li, Amir Rahmati, Chaowei
  Xiao, Atul Prakash, Tadayoshi Kohno, and Dawn Song.
\newblock Robust physical-world attacks on deep learning visual classification.
\newblock In {\em Proceedings of the IEEE conference on computer vision and
  pattern recognition}, pages 1625--1634, 2018.

\bibitem{goodfellow2014generative}
Ian Goodfellow, Jean Pouget-Abadie, Mehdi Mirza, Bing Xu, David Warde-Farley,
  Sherjil Ozair, Aaron Courville, and Yoshua Bengio.
\newblock Generative adversarial nets.
\newblock {\em Advances in neural information processing systems}, 27, 2014.

\bibitem{goodfellow2014explaining}
Ian~J Goodfellow, Jonathon Shlens, and Christian Szegedy.
\newblock Explaining and harnessing adversarial examples.
\newblock {\em arXiv preprint arXiv:1412.6572}, 2014.

\bibitem{he2016deep}
Kaiming He, Xiangyu Zhang, Shaoqing Ren, and Jian Sun.
\newblock Deep residual learning for image recognition.
\newblock In {\em Proceedings of the IEEE conference on computer vision and
  pattern recognition}, pages 770--778, 2016.

\bibitem{hill2020stochastic}
Mitch Hill, Jonathan Mitchell, and Song-Chun Zhu.
\newblock Stochastic security: Adversarial defense using long-run dynamics of
  energy-based models.
\newblock {\em arXiv preprint arXiv:2005.13525}, 2020.

\bibitem{huang2023fast}
Zhichao Huang, Yanbo Fan, Chen Liu, Weizhong Zhang, Yong Zhang, Mathieu
  Salzmann, Sabine S{\"u}sstrunk, and Jue Wang.
\newblock Fast adversarial training with adaptive step size.
\newblock {\em IEEE Transactions on Image Processing}, 2023.

\bibitem{hwang2019puvae}
Uiwon Hwang, Jaewoo Park, Hyemi Jang, Sungroh Yoon, and Nam~Ik Cho.
\newblock Puvae: A variational autoencoder to purify adversarial examples.
\newblock {\em IEEE Access}, 7:126582--126593, 2019.

\bibitem{jia2023improving}
Xiaojun Jia, Yong Zhang, Xingxing Wei, Baoyuan Wu, Ke Ma, Jue Wang, and
  Xiaochun Cao~Sr.
\newblock Improving fast adversarial training with prior-guided knowledge.
\newblock {\em arXiv preprint arXiv:2304.00202}, 2023.

\bibitem{kim2021understanding}
Hoki Kim, Woojin Lee, and Jaewook Lee.
\newblock Understanding catastrophic overfitting in single-step adversarial
  training.
\newblock In {\em Proceedings of the AAAI Conference on Artificial
  Intelligence}, volume~35, pages 8119--8127, 2021.

\bibitem{kingma2013auto}
Diederik~P Kingma and Max Welling.
\newblock Auto-encoding variational bayes.
\newblock {\em arXiv preprint arXiv:1312.6114}, 2013.

\bibitem{krizhevsky2009learning}
Alex Krizhevsky, Geoffrey Hinton, et~al.
\newblock Learning multiple layers of features from tiny images.
\newblock 2009.

\bibitem{kurakin2016adversarial}
Alexey Kurakin, Ian Goodfellow, and Samy Bengio.
\newblock Adversarial machine learning at scale.
\newblock {\em arXiv preprint arXiv:1611.01236}, 2016.

\bibitem{le2015tiny}
Ya Le and Xuan Yang.
\newblock Tiny imagenet visual recognition challenge.
\newblock {\em CS 231N}, 7(7):3, 2015.

\bibitem{lecun1998gradient}
Yann LeCun, L{\'e}on Bottou, Yoshua Bengio, and Patrick Haffner.
\newblock Gradient-based learning applied to document recognition.
\newblock {\em Proceedings of the IEEE}, 86(11):2278--2324, 1998.

\bibitem{li2020towards}
Bai Li, Shiqi Wang, Suman Jana, and Lawrence Carin.
\newblock Towards understanding fast adversarial training.
\newblock {\em arXiv preprint arXiv:2006.03089}, 2020.

\bibitem{li2022subspace}
Tao Li, Yingwen Wu, Sizhe Chen, Kun Fang, and Xiaolin Huang.
\newblock Subspace adversarial training.
\newblock In {\em Proceedings of the IEEE/CVF Conference on Computer Vision and
  Pattern Recognition}, pages 13409--13418, 2022.

\bibitem{liao2018defense}
Fangzhou Liao, Ming Liang, Yinpeng Dong, Tianyu Pang, Xiaolin Hu, and Jun Zhu.
\newblock Defense against adversarial attacks using high-level representation
  guided denoiser.
\newblock In {\em Proceedings of the IEEE conference on computer vision and
  pattern recognition}, pages 1778--1787, 2018.

\bibitem{lin2020dual}
Wei-An Lin, Chun~Pong Lau, Alexander Levine, Rama Chellappa, and Soheil Feizi.
\newblock Dual manifold adversarial robustness: Defense against lp and non-lp
  adversarial attacks.
\newblock {\em Advances in Neural Information Processing Systems},
  33:3487--3498, 2020.

\bibitem{madry2017towards}
Aleksander Madry, Aleksandar Makelov, Ludwig Schmidt, Dimitris Tsipras, and
  Adrian Vladu.
\newblock Towards deep learning models resistant to adversarial attacks.
\newblock {\em arXiv preprint arXiv:1706.06083}, 2017.

\bibitem{moosavi2016deepfool}
Seyed-Mohsen Moosavi-Dezfooli, Alhussein Fawzi, and Pascal Frossard.
\newblock Deepfool: a simple and accurate method to fool deep neural networks.
\newblock In {\em Proceedings of the IEEE conference on computer vision and
  pattern recognition}, pages 2574--2582, 2016.

\bibitem{netzer2011reading}
Yuval Netzer, Tao Wang, Adam Coates, Alessandro Bissacco, Bo Wu, and Andrew~Y
  Ng.
\newblock Reading digits in natural images with unsupervised feature learning.
\newblock 2011.

\bibitem{nie2022diffusion}
Weili Nie, Brandon Guo, Yujia Huang, Chaowei Xiao, Arash Vahdat, and Anima
  Anandkumar.
\newblock Diffusion models for adversarial purification.
\newblock {\em arXiv preprint arXiv:2205.07460}, 2022.

\bibitem{robbins1951stochastic}
Herbert Robbins and Sutton Monro.
\newblock A stochastic approximation method.
\newblock {\em The annals of mathematical statistics}, pages 400--407, 1951.

\bibitem{rony2019decoupling}
J{\'e}r{\^o}me Rony, Luiz~G Hafemann, Luiz~S Oliveira, Ismail~Ben Ayed, Robert
  Sabourin, and Eric Granger.
\newblock Decoupling direction and norm for efficient gradient-based l2
  adversarial attacks and defenses.
\newblock In {\em Proceedings of the IEEE/CVF Conference on Computer Vision and
  Pattern Recognition}, pages 4322--4330, 2019.

\bibitem{ruder2016overview}
Sebastian Ruder.
\newblock An overview of gradient descent optimization algorithms.
\newblock {\em arXiv preprint arXiv:1609.04747}, 2016.

\bibitem{samangouei2018defense}
Pouya Samangouei, Maya Kabkab, and Rama Chellappa.
\newblock Defense-gan: Protecting classifiers against adversarial attacks using
  generative models.
\newblock {\em arXiv preprint arXiv:1805.06605}, 2018.

\bibitem{selvaraju2017grad}
Ramprasaath~R Selvaraju, Michael Cogswell, Abhishek Das, Ramakrishna Vedantam,
  Devi Parikh, and Dhruv Batra.
\newblock Grad-cam: Visual explanations from deep networks via gradient-based
  localization.
\newblock In {\em Proceedings of the IEEE international conference on computer
  vision}, pages 618--626, 2017.

\bibitem{shi2021online}
Changhao Shi, Chester Holtz, and Gal Mishne.
\newblock Online adversarial purification based on self-supervision.
\newblock {\em arXiv preprint arXiv:2101.09387}, 2021.

\bibitem{simonyan2014very}
Karen Simonyan and Andrew Zisserman.
\newblock Very deep convolutional networks for large-scale image recognition.
\newblock {\em arXiv preprint arXiv:1409.1556}, 2014.

\bibitem{sriramanan2020guided}
Gaurang Sriramanan, Sravanti Addepalli, Arya Baburaj, et~al.
\newblock Guided adversarial attack for evaluating and enhancing adversarial
  defenses.
\newblock {\em Advances in Neural Information Processing Systems},
  33:20297--20308, 2020.

\bibitem{sriramanan2021towards}
Gaurang Sriramanan, Sravanti Addepalli, Arya Baburaj, et~al.
\newblock Towards efficient and effective adversarial training.
\newblock {\em Advances in Neural Information Processing Systems},
  34:11821--11833, 2021.

\bibitem{szegedy2013intriguing}
Christian Szegedy, Wojciech Zaremba, Ilya Sutskever, Joan Bruna, Dumitru Erhan,
  Ian Goodfellow, and Rob Fergus.
\newblock Intriguing properties of neural networks.
\newblock {\em arXiv preprint arXiv:1312.6199}, 2013.

\bibitem{tsipras2018robustness}
Dimitris Tsipras, Shibani Santurkar, Logan Engstrom, Alexander Turner, and
  Aleksander Madry.
\newblock Robustness may be at odds with accuracy.
\newblock {\em arXiv preprint arXiv:1805.12152}, 2018.

\bibitem{van2008visualizing}
Laurens Van~der Maaten and Geoffrey Hinton.
\newblock Visualizing data using t-sne.
\newblock {\em Journal of machine learning research}, 9(11), 2008.

\bibitem{vivek2019regularizer}
BS Vivek, Arya Baburaj, and R~Venkatesh Babu.
\newblock Regularizer to mitigate gradient masking effect during single-step
  adversarial training.
\newblock In {\em CVPR Workshops}, pages 66--73, 2019.

\bibitem{wang2022guided}
Jinyi Wang, Zhaoyang Lyu, Dahua Lin, Bo Dai, and Hongfei Fu.
\newblock Guided diffusion model for adversarial purification.
\newblock {\em arXiv preprint arXiv:2205.14969}, 2022.

\bibitem{wang2019improving}
Yisen Wang, Difan Zou, Jinfeng Yi, James Bailey, Xingjun Ma, and Quanquan Gu.
\newblock Improving adversarial robustness requires revisiting misclassified
  examples.
\newblock In {\em International conference on learning representations}, 2019.

\bibitem{wang2023better}
Zekai Wang, Tianyu Pang, Chao Du, Min Lin, Weiwei Liu, and Shuicheng Yan.
\newblock Better diffusion models further improve adversarial training.
\newblock {\em arXiv preprint arXiv:2302.04638}, 2023.

\bibitem{wei2023cfa}
Zeming Wei, Yifei Wang, Yiwen Guo, and Yisen Wang.
\newblock Cfa: Class-wise calibrated fair adversarial training.
\newblock In {\em Proceedings of the IEEE/CVF Conference on Computer Vision and
  Pattern Recognition}, pages 8193--8201, 2023.

\bibitem{wong2020fast}
Eric Wong, Leslie Rice, and J~Zico Kolter.
\newblock Fast is better than free: Revisiting adversarial training.
\newblock {\em arXiv preprint arXiv:2001.03994}, 2020.

\bibitem{wu2020adversarial}
Dongxian Wu, Shu-Tao Xia, and Yisen Wang.
\newblock Adversarial weight perturbation helps robust generalization.
\newblock {\em Advances in Neural Information Processing Systems},
  33:2958--2969, 2020.

\bibitem{wu2020stronger}
Kaiwen Wu, Allen Wang, and Yaoliang Yu.
\newblock Stronger and faster wasserstein adversarial attacks.
\newblock In {\em International Conference on Machine Learning}, pages
  10377--10387. PMLR, 2020.

\bibitem{xiao2018spatially}
Chaowei Xiao, Jun-Yan Zhu, Bo Li, Warren He, Mingyan Liu, and Dawn Song.
\newblock Spatially transformed adversarial examples.
\newblock {\em arXiv preprint arXiv:1801.02612}, 2018.

\bibitem{xie2019improving}
Cihang Xie, Zhishuai Zhang, Yuyin Zhou, Song Bai, Jianyu Wang, Zhou Ren, and
  Alan~L Yuille.
\newblock Improving transferability of adversarial examples with input
  diversity.
\newblock In {\em Proceedings of the IEEE/CVF conference on computer vision and
  pattern recognition}, pages 2730--2739, 2019.

\bibitem{xu2023exploring}
Yuancheng Xu, Yanchao Sun, Micah Goldblum, Tom Goldstein, and Furong Huang.
\newblock Exploring and exploiting decision boundary dynamics for adversarial
  robustness.
\newblock {\em arXiv preprint arXiv:2302.03015}, 2023.

\bibitem{yang2021class}
Kaiwen Yang, Tianyi Zhou, Yonggang Zhang, Xinmei Tian, and Dacheng Tao.
\newblock Class-disentanglement and applications in adversarial detection and
  defense.
\newblock {\em Advances in Neural Information Processing Systems},
  34:16051--16063, 2021.

\bibitem{yang2019me}
Yuzhe Yang, Guo Zhang, Dina Katabi, and Zhi Xu.
\newblock Me-net: Towards effective adversarial robustness with matrix
  estimation.
\newblock {\em arXiv preprint arXiv:1905.11971}, 2019.

\bibitem{yuan2020ensemble}
Jianhe Yuan and Zhihai He.
\newblock Ensemble generative cleaning with feedback loops for defending
  adversarial attacks.
\newblock In {\em Proceedings of the IEEE/CVF Conference on Computer Vision and
  Pattern Recognition}, pages 581--590, 2020.

\bibitem{zagoruyko2016wide}
Sergey Zagoruyko and Nikos Komodakis.
\newblock Wide residual networks.
\newblock {\em arXiv preprint arXiv:1605.07146}, 2016.

\bibitem{zhang2019theoretically}
Hongyang Zhang, Yaodong Yu, Jiantao Jiao, Eric Xing, Laurent El~Ghaoui, and
  Michael Jordan.
\newblock Theoretically principled trade-off between robustness and accuracy.
\newblock In {\em International conference on machine learning}, pages
  7472--7482. PMLR, 2019.

\bibitem{zhang2020attacks}
Jingfeng Zhang, Xilie Xu, Bo Han, Gang Niu, Lizhen Cui, Masashi Sugiyama, and
  Mohan Kankanhalli.
\newblock Attacks which do not kill training make adversarial learning
  stronger.
\newblock In {\em International conference on machine learning}, pages
  11278--11287. PMLR, 2020.

\bibitem{zhou2021towards}
Dawei Zhou, Tongliang Liu, Bo Han, Nannan Wang, Chunlei Peng, and Xinbo Gao.
\newblock Towards defending against adversarial examples via attack-invariant
  features.
\newblock In {\em International Conference on Machine Learning}, pages
  12835--12845. PMLR, 2021.

\bibitem{zhou2021removing}
Dawei Zhou, Nannan Wang, Chunlei Peng, Xinbo Gao, Xiaoyu Wang, Jun Yu, and
  Tongliang Liu.
\newblock Removing adversarial noise in class activation feature space.
\newblock In {\em Proceedings of the IEEE/CVF International Conference on
  Computer Vision}, pages 7878--7887, 2021.

\end{thebibliography}
}

\end{document}